\documentclass[preprint,12pt]{elsarticle}

\usepackage[utf8]{inputenc} 
\usepackage[T1]{fontenc}    
\usepackage{hyperref}       
\usepackage{url}            
\usepackage{booktabs}       
\usepackage{amsfonts}       
\usepackage{nicefrac}       
\usepackage{microtype}      
\usepackage{xcolor}         
\usepackage{graphicx}
\usepackage{subcaption}
\usepackage{siunitx}
\usepackage{float}
\usepackage{pdfpages}
\usepackage{svg}
\usepackage{tikz}
\usetikzlibrary{arrows.meta, positioning}
\usepackage{makecell}
\usepackage{multirow}

\setcellgapes{2pt}
\makegapedcells

\journal{Neural Networks}

\begin{document}

\begin{frontmatter}

\title{Parallelizing Node-Level Explainability in Graph Neural Networks}


\author[inst1,inst2]{Oscar Llorente\corref{cor1}}
\ead{oscar.llorente.gonzalez@ericsson.com}

\author[inst2]{Jaime Boal}
\ead{jaime.boal@iit.comillas.edu}

\author[inst2]{Eugenio F. Sánchez-Úbeda}
\ead{eugenio.sanchez@iit.comillas.edu}

\author[inst1]{Antonio Diaz-Cano}
\ead{antonio.jesus.diaz.cano@ericsson.com}

\author[inst1]{Miguel Familiar}
\ead{miguel.familiar-cabero@ericsson.com}

\affiliation[inst1]{organization={Ericsson Cognitive Labs},
            addressline={Retama Ed 1 Torre Suecia}, 
            city={Madrid},
            postcode={28045}, 
            state={Madrid},
            country={Spain}}

\affiliation[inst2]{organization={Institute for Research in Technology~(IIT), ICAI School of Engineering, Comillas Pontifical University},
            addressline={Rey Francisco, 4}, 
            city={Madrid},
            postcode={28008}, 
            state={Madrid},
            country={Spain}}

\begin{abstract}
  Graph Neural Networks~(GNNs) have demonstrated remarkable performance in a wide range of tasks, such as node classification, link prediction, and graph classification, by exploiting the structural information in graph-structured data. However, in node classification, computing node-level explainability becomes extremely time-consuming as the size of the graph increases, while batching strategies often degrade explanation quality. This paper introduces a novel approach to parallelizing node-level explainability in GNNs through graph partitioning. By decomposing the graph into disjoint subgraphs, we enable parallel computation of explainability for node neighbors, significantly improving the scalability and efficiency without affecting the correctness of the results, provided sufficient memory is available. For scenarios where memory is limited, we further propose a dropout-based reconstruction mechanism that offers a controllable trade-off between memory usage and explanation fidelity. Experimental results on real-world datasets demonstrate substantial speedups, enabling scalable and transparent explainability for large-scale GNN models~\footnote{The source code is available at \url{https://github.com/EricssonResearch/parallel-xai-gnn}}.
\end{abstract}


\end{frontmatter}

\section{Introduction}

Graphs are common structures for representing relationships in data across several domains. Leveraging the potential of these complex relational structures, Graph Neural Networks~(GNNs) have transformed the way such data is processed and analyzed in domains such as medicine~\cite{ABADAL2025106792}, chemistry~\cite{reiser_graph_2022}, telecommunications~\cite{9252917}, or social networks~\cite{10.1145/3308558.3313488}. However, their inherent complexity and black-box nature pose significant challenges, especially in applications requiring trust and interpretability~\cite{WANG2026108448}. The growing need to understand and interpret GNN decisions has driven the development of various explainability methods. Although these methods can provide meaningful insights, they are inherently sequential and computationally intensive, which often leads to scalability challenges, as discussed later; this makes them impractical for production environments and real-time applications.

Broadly, interpretability can be pursued in two ways: designing models that are inherently transparent or developing techniques that explain black-box deep learning models. Following the terminology in~\cite{GNNBook}, the former is referred to as \emph{interpretable modelling}, and the latter as \emph{post-hoc explainability}, which is the setting addressed in this work.

Post-hoc explainability for deep neural networks has been an active research area in the past decade~\cite{simonyanDeepConvolutionalNetworks2014a, SpringenbergDBR14, 7780688, selvarajuGradCAMVisualExplanations2017, DBLP:conf/bmvc/PetsiukDS18, Kim2019WhyAS}. Early work focused on computer vision~\cite{simonyanDeepConvolutionalNetworks2014a, selvarajuGradCAMVisualExplanations2017, zhouLearningDeepFeatures2016}, and the field has since expanded to natural language processing~\cite{danilevsky} and GNNs~\cite{li2022explainabilitygraphneuralnetworks}. Within the literature, two main goals are commonly distinguished~\cite{9875989}: \emph{instance-level explanations}, which seek to clarify why a specific prediction has been made, and \emph{model-level explanations}, which characterize the general behavior of the model. Our work falls into the instance-level category.

Gradient-based techniques such as Saliency Maps~\cite{simonyanDeepConvolutionalNetworks2014a}, Guided Backpropagation~\cite{SpringenbergDBR14}, Smoothgrad~\cite{Smilkov2017SmoothGradRN}, or RectGrad~\cite{Kim2019WhyAS} remain among the most widely used approaches. However, these methods can produce unreliable explanations, sometimes assigning high importance to features that are not truly influential~\cite{sanity}. This limitation has motivated research on evaluating explainability methods themselves~\cite{DBLP:conf/bmvc/PetsiukDS18,benchmarkxai}, introducing metrics based on perturbing or removing (ablating) parts of the inputs according to their assigned importance scores.

In GNNs, each node representation is updated by aggregating information from its neighbors and combining it with its own features, a process known as \emph{message-passing}~\cite{passing}. Therefore, in a $k$-layer GNN the predictions will consider nodes up to $k$-hops away from the target node. This approach has made GNNs successful lately, but introduces some challenges for explainability. In terms of prediction, GNN tasks are commonly grouped into graph-level, node-level, and link-level problems. The explainability requirements of these three problems differ substantially:

\begin{itemize}
    \item For \emph{graph-level} prediction, explainability typically focuses on identifying a subgraph (i.e., a subset of the most influential nodes) that preserves the original prediction of the model~\cite{gnnexplainer, pgexplainer}.
    \item For \emph{node-level} prediction, one may aim to assess the contribution of each feature to the predicted outcome of a node~\cite{baldasarre}, or assign a global importance score to every node~\cite{evaluatinggnnxai}. In this setting, the goal is to explain how much each node contributes to the prediction of a specific target node, yielding one explanation per node prediction.
    \item For \emph{link-level} prediction, the objective is to quantify the importance of features or neighboring nodes~\cite{xailink}.
\end{itemize}

Following the convention in the literature~\cite{gnnexplainer, pgexplainer, evaluatinggnnxai}, we use the term \emph{importance} to denote the extent to which a feature or a neighbor node influences the prediction made by a GNN. This notion is inherently model-specific, as it reflects what the GNN has learned. For example, in graph classification, a GNN may rely more heavily on certain features, whose influence on the predictions defines their importance. Consequently, another model may emphasize other features, indicating that it has learned different underlying patterns.

\begin{figure}[ht]
  \centering
  \begin{subfigure}{0.49\linewidth}
      \centering
      \includegraphics[width=\linewidth, height=0.8\linewidth]{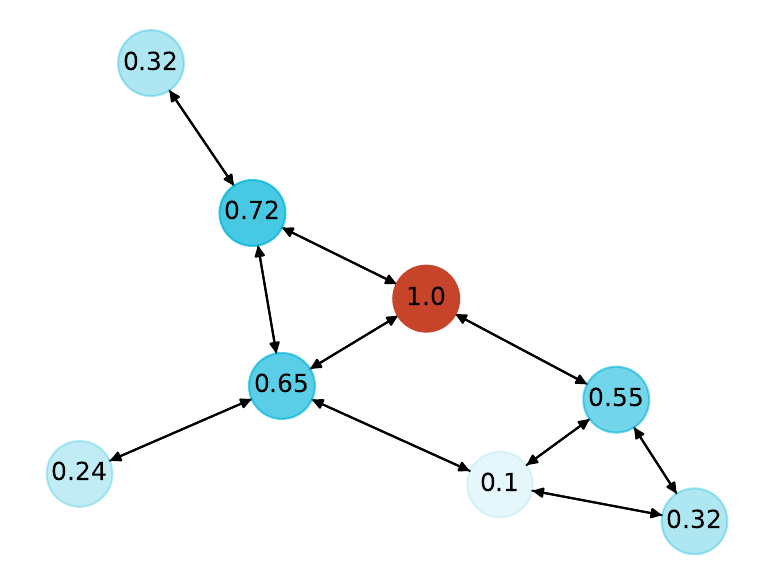}
      \caption{Saliency Map}
      \label{fig: saliency map}
  \end{subfigure}
  \begin{subfigure}{0.49\linewidth}
    \centering
    \includegraphics[width=\linewidth, height=0.8\linewidth]{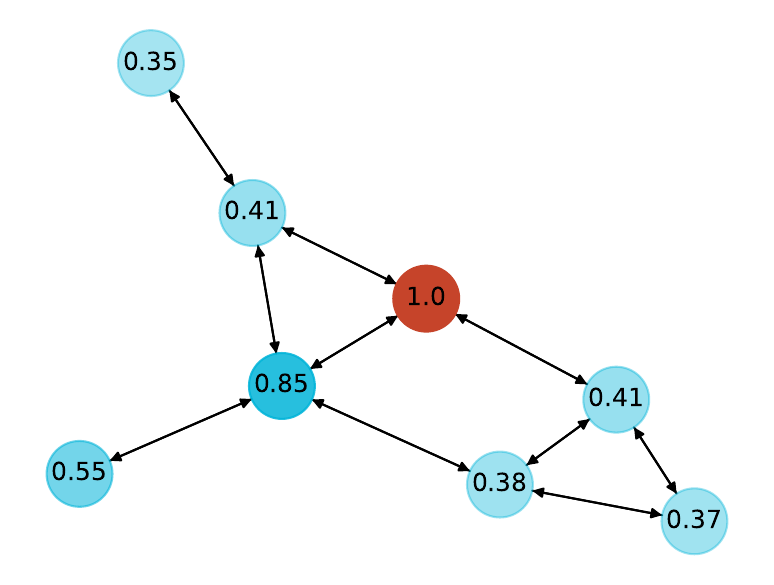}
    \caption{GNNExplainer}
    \label{fig: gnnexplainer}
  \end{subfigure}
  \caption{Example of node importance scores computed using two explainability techniques: (a) Saliency Map~\cite{simonyanDeepConvolutionalNetworks2014a}, (b) GNNExplainer~\cite{gnnexplainer}. Scores are shown for node classification on the Cora dataset using a Graph Convolutional Network~(GCN)~\cite{gcn}. The red node is the target being classified, and the surrounding nodes are the neighbors considered. Importance values are in the $\left[0,1\right]$ range, with 1 indicating maximum importance.}
  \label{fig: explainability techniques}
\end{figure}

This work focuses on the node-level task of quantifying node importance. This problem is particularly relevant for industrial applications (e.g.,~telecommunications~\cite{varela} or power~\cite{elec} networks), where predictions may be required for every node and explanations must be provided individually. Following~\cite{evaluatinggnnxai}, we compute a single explainability score per node, assigning an importance value to every neighbor and to the node itself (\autoref{fig: explainability techniques}). Nevertheless, our methodology can be directly applied to scenarios where fine-grained feature-level scores are desirable.

Computing explainability scores for all node-level predictions simultaneously is not feasible and must be generated sequentially. This is because when multiple target nodes share a neighbor, its explainability score becomes a combined result of the contributions of the different predictions, corrupting the results. In production settings, this sequential computation introduces significant overhead, making the process extremely slow when both the graph and the number of predictions to explain are large. This paper proposes a technique to overcome this limitation, enabling the parallel computation of node-level explainability in GNNs. To the best of our knowledge, this is the first work to tackle this problem in the literature; previous studies, such as~\cite{evaluatinggnnxai} and~\cite{taxonomic}, rely on sequential execution.

In summary, this paper (i)~identifies the ``gradient-mixing'' bottleneck that prevents batch execution for node-level gradient-based GNN explainability; (ii)~introduces a general parallelization framework based on graph partitioning that accelerates computation without altering explanation quality; and (iii)~introduces a memory–accuracy trade-off mechanism through dropout reconstruction to mitigate the memory overhead of the proposed method. Together, these contributions enable practical, production-scale explainability for large GNNs.

The remainder of this paper is organized as follows. Section~\ref{sec: parallel} analyzes why computing node-level explanations in GNNs is inherently time-consuming, motivating the need for parallelization. Section~\ref{sec: methodology} describes the proposed methodology based on graph partitioning and reconstruction. Section~\ref{sec: theoretical} provides a theoretical analysis of the computational complexity and expected performance gains. Section~\ref{sec: results} reports experimental results and discusses the trade-offs observed across datasets and explainability methods. Section~\ref{sec: conclusion} concludes the paper and outlines future research directions. Finally, \ref{sec: complete results} presents the complete set of experimental results for reference.

\section{Why is computing explanations for GNNs time-consuming?}
\label{sec: parallel}

This section provides a detailed discussion of the challenges in parallelizing GNN explainability. We first focus on gradient-based techniques, which typically achieve the best results~\cite{reiser_graph_2022,evaluatinggnnxai,pope}, and then describe how similar issues affect other types of explainability methods.

Gradient-based methods cannot be executed in parallel due to what we refer to as the ``gradient-mixing problem'', which forces sequential computation for each node. Other techniques that are not strictly classified as gradient-based but rely on gradients in their computations, such as GNNExplainer~\cite{gnnexplainer}, are also affected. 

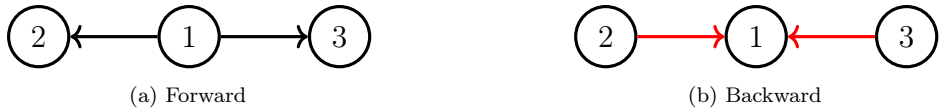
\begin{figure}[ht]
    \centering
    \begin{subfigure}[b]{0.45\textwidth}
        \centering
        \begin{tikzpicture}[->, very thick, scale=2, every node/.style={circle,draw, minimum size=0.8cm}]
            \node (1) at (0,0) {1};
            \node (2) at (-1,0) {2};
            \node (3) at (1,0) {3};
            \draw[->] (1) -- (2);
            \draw[->] (1) -- (3);
        \end{tikzpicture}
        \caption{Forward}
        \label{fig: forward}
    \end{subfigure}
    \hfill
    \begin{subfigure}[b]{0.45\textwidth}
        \centering
        \begin{tikzpicture}[->, very thick, scale=2, every node/.style={circle,draw, minimum size=0.8cm}]
            \node (1) at (0,0) {1};
            \node (2) at (-1,0) {2};
            \node (3) at (1,0) {3};
            \draw[->, red] (2) -- (1);
            \draw[->, red] (3) -- (1);
        \end{tikzpicture}
        \caption{Backward}
    \end{subfigure}
    \caption{Forward and backward passes of a 1-hop GNN. (a) Black lines represent the forward flow of information; (b) red lines indicate the backward flow during gradient computation.}
    \label{fig: forward and backward}
\end{figure}

To illustrate the problem, consider a graph where a node serves as a neighbor to two other nodes~(\autoref{fig: forward and backward}). Let this node be node~1, which is a neighbor of nodes~2 and~3. Using a 1-hop GNN, both~2 and~3 aggregate information from node 1 during the forward pass to make their predictions. When computing explainability scores, the gradient flows back from node~2 to node~1, and from node~3 to node~1. These two contributions are merged, producing a single overall importance score for node 1 rather than a distinct score for each prediction. This behavior occurs for any number of hops whenever nodes share a common neighbor, preventing the extraction of per-prediction explainability information.

A similar situation occurs for perturbation-based techniques, such as~\cite{zorro}. If a node is a neighbor of multiple other nodes, modifying its value affects all its neighbors simultaneously. Consequently, if the explainability score depends on the impact of the changed features on the neighbors, the result will again be a combined score rather than one per target node. For example, if node~1 in~\autoref{fig: forward} is perturbed, the predictions of nodes~2 and~3 will both change. Since perturbation-based explainability techniques rely on the magnitude of these changes, node~1 will receive a single combined importance score instead of separate scores corresponding to nodes~2 and~3, even though the goal is to explain both predictions individually. 

It is possible to adapt perturbation-based methods to account for this issue, providing an alternative potential path for parallelization. In contrast, gradient-based methods inherently produce combined information in the scenario shown in~\autoref{fig: forward and backward}, since their computations depend on the implementation of automatic differentiation (autodiff)~\cite{autodiff}. The method presented in this paper is a general methodology for parallelizing GNN explainability compatible with all techniques. Developing a specialized approach for perturbation-based methods could be an interesting direction for future work but is beyond the scope of this study.

Other methods that do not belong to the gradient-based or perturbation-based categories, like PGExplainer~\cite{pgexplainer}, have a similar underlying issue to that encountered with perturbation-based techniques. While it would be possible to modify the current implementations to address this problem, such solutions would be method-specific rather than general. 

\section{Methodology}
\label{sec: methodology}

\begin{figure}[t]
  \centering
  \includegraphics[width=\textwidth]{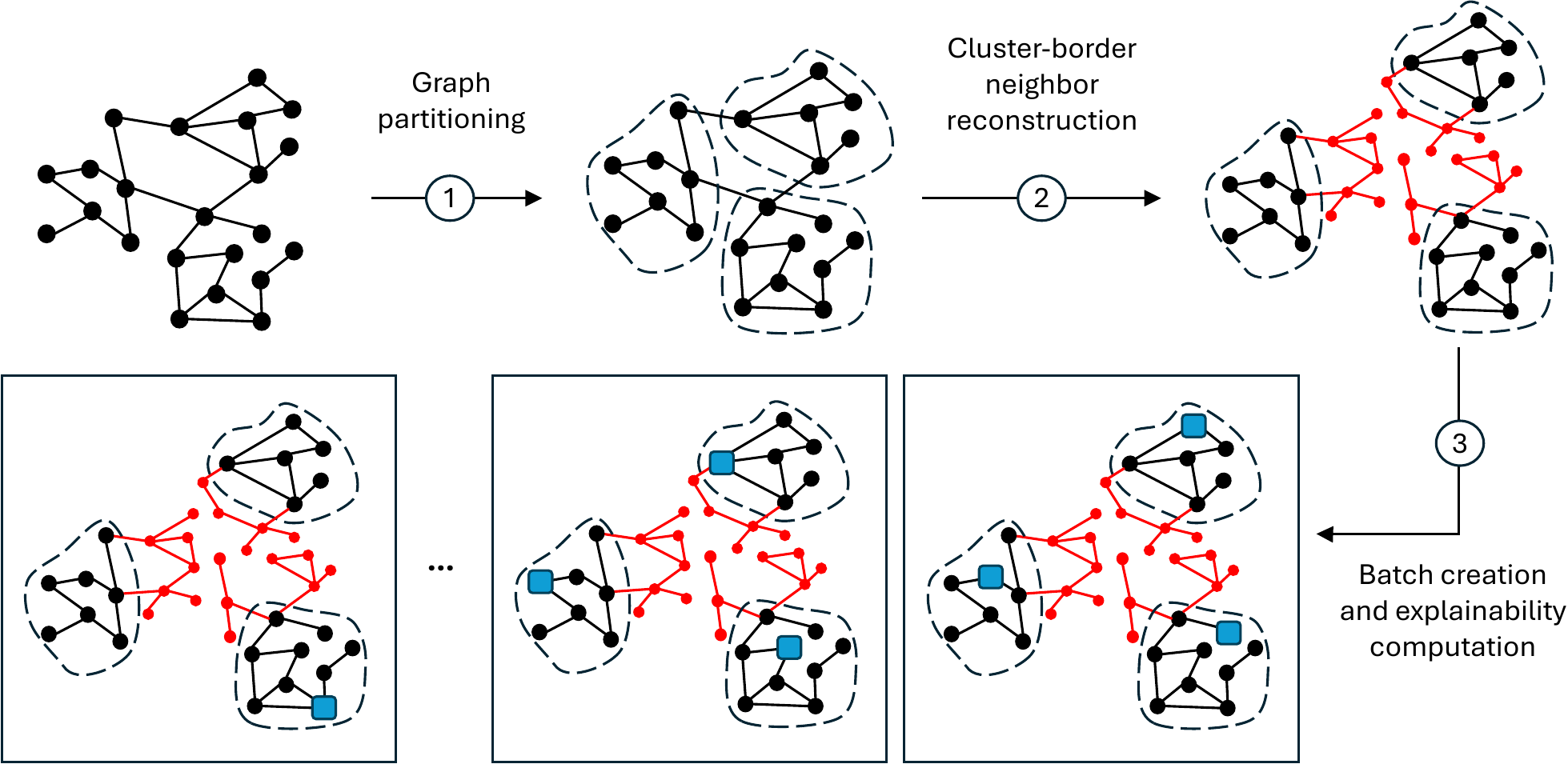}
  \label{fig: overview_methodology}
  \caption{Overview of the proposed methodology. Red nodes and edges represent the reconstructed neighbors, while blue nodes denote, for each batch, the selected nodes from each cluster for which explainability is executed in parallel.}
  \label{fig: overview methodology}
\end{figure}

When applying node-level explainability in GNNs, the results may be contaminated whenever predictions share common neighbors. Our solution is to construct batches of independent nodes (i.e.,~sets of nodes that do not influence one another through shared neighbors). Explanations can then be computed in parallel within each batch, ensuring that no interference occurs. The proposed methodology consists of the following steps, executed sequentially (\autoref{fig: overview methodology}):

\begin{enumerate}
    \item Graph partitioning. The graph is partitioned into a given number of clusters, ideally minimizing the number of broken edges.
    \item Reconstruction of broken cluster-border neighbors. Neighboring nodes across cluster borders are reconstructed based on the number of hops required for the correct evaluation of the GNN. In the illustrative example of \autoref{fig: overview methodology}, 2-hop neighbors are employed.
    \item Batch creation and explainability computation. Based on the reconstructed clusters, batches of nodes are created by selecting one node per cluster, enabling the parallel computation of explainability scores. The number of clusters defines the batch size of most parallel executions; however, due to imbalances in cluster sizes, the final iterations may involve smaller batches, as some clusters contain fewer nodes than others. In the example of \autoref{fig: overview methodology}, since three clusters are used, each batch ideally consists of three nodes. However, because the clusters contain different numbers of nodes, complete 3-tuple batches are not possible in all cases.
\end{enumerate}

\subsection{Graph partitioning}

\begin{figure}
    \centering
    \begin{subfigure}[b]{0.45\textwidth}
        \centering
        \begin{tikzpicture}[
                thick, 
                roundnode/.style={circle, draw=black, very thick, minimum size=7mm},
            ]
            \node[roundnode] (one) {1};
            \node[roundnode] (two) [below= of one] {2};
            \node[roundnode] (three) [right=of one] {3};
            \node[roundnode] (four) [below=of three] {4};
            \node[roundnode] (five) [above=of three] {5};
            \node[roundnode] (six) [right=of five] {6};
            \node[roundnode] (seven) [below=of six] {7};
            \node[roundnode] (eight) [right=of seven] {8};
            \node[roundnode] (nine) [below=of eight] {9};
            
            \draw[->] (two.north) -- (one.south);
            \draw[->] (one.east) -- (three.west);
            \draw[->] (four.north) -- (three.south);
            \draw[->] (three.north) -- (five.south);
            \draw[->] (five.east) -- (six.west);
            \draw[->] (seven.north) -- (six.south);
            \draw[->] (eight.west) -- (seven.east);
            \draw[->] (nine.north) -- (eight.south);
        \end{tikzpicture}
        \caption{Original graph}
        \label{fig: original graph}
    \end{subfigure}
    \hfill
    \begin{subfigure}[b]{0.45\textwidth}
        \centering
        \begin{tikzpicture}[
                thick,
                roundnode/.style={circle, draw=black, very thick, minimum size=7mm},
                bluenode/.style={circle, draw=teal, very thick, minimum size=7mm},
                magentanode/.style={circle, draw=olive, very thick, minimum size=7mm},
                greennode/.style={circle, draw=purple, very thick, minimum size=7mm},
            ]
            \node[bluenode] (one) {1};
            \node[bluenode] (two) [below= of one] {2};
            \node[bluenode] (three) [right=of one] {3};
            \node[bluenode] (four) [below=of three] {4};
            \node[magentanode] (five) [above=of three] {5};
            \node[magentanode] (six) [right=of five] {6};
            \node[greennode] (seven) [below=of six] {7};
            \node[greennode] (eight) [right=of seven] {8};
            \node[greennode] (nine) [below=of eight] {9};
            
            \draw[->] (two.north) -- (one.south);
            \draw[->] (one.east) -- (three.west);
            \draw[->] (four.north) -- (three.south);
            \draw[->] (five.east) -- (six.west);
            \draw[->] (eight.west) -- (seven.east);
            \draw[->] (nine.north) -- (eight.south);
        \end{tikzpicture}
        \caption{Clusters}
        \label{fig: clusters}
    \end{subfigure}
    \caption{Example of graph partitioning into clusters, first step of the proposed parallelization process. (a) Original graph; (b) cluster division of the graph, where each color represents a different cluster. The edges 3-5 and 7-6 have been broken due to the cluster creation process.}
    \label{fig: division}
\end{figure}
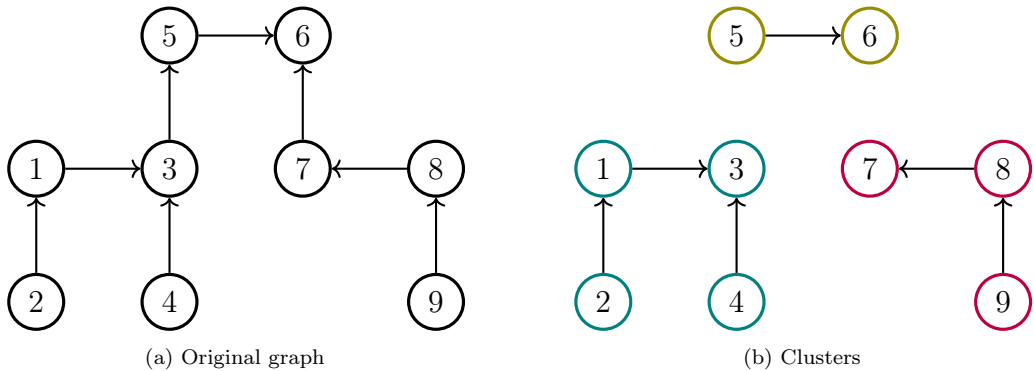

First, we divide the graph into independent clusters (see example in \autoref{fig: division}), so that after one node per cluster is selected, gradient-mixing is avoided. Creating clusters will break some graph edges, so the predictions and the explainability will not be the same for the nodes close to the cluster border. More precisely, in a $k$-hop GNN, predictions will be affected in nodes that are strictly less than $k$ hops apart from the border of the cluster. To minimize this issue, we propose using the METIS algorithm~\cite{doi:10.1137/S1064827595287997}, which creates partitions that minimize the number of broken edges.

In METIS, as in most clustering and graph-partitioning algorithms, the number of clusters~$c$ is a hyperparameter that must be specified in advance. The algorithm does not determine an optimal value of~$c$, it simply generates a balanced partition while minimizing edge cuts for the user-defined number of clusters. In our case, since larger values of~$c$ enable greater parallelization, it should be chosen to reach a reasonable trade-off between total execution time and memory usage, which increases with the number of nodes and edges that need to be reconstructed at cluster borders.

\subsection{Reconstruction of cluster-border neighbors}
\label{sec: reconstruction}

The second step reconstructs the cluster borders that were broken during partitioning, ensuring that the explainability results remain consistent with those obtained without clustering. We propose two variants: \emph{full reconstruction} and \emph{dropout reconstruction}.  

\subsubsection{Full reconstruction}

\begin{figure}
    \centering
    \begin{subfigure}[b]{0.45\textwidth}
        \centering
        \begin{tikzpicture}[
                thick,
                bluenode/.style={circle, draw=teal, very thick, minimum size=7mm},
                magentanode/.style={circle, draw=olive, very thick, minimum size=7mm},
                greennode/.style={circle, draw=purple, very thick, minimum size=7mm},
                graynode/.style={circle, draw=darkgray, fill=gray!30, very thick, minimum size=7mm},
            ]
            \node[bluenode] (one) {1};
            \node[bluenode] (two) [below= of one] {2};
            \node[bluenode] (three) [right=of one] {3};
            \node[bluenode] (four) [below=of three] {4};
            \node[magentanode] (five) [above=of three] {5};
            \node[magentanode] (six) [right=of five] {6};
            \node[greennode] (seven) [below=of six] {7};
            \node[greennode] (eight) [right=of seven] {8};
            \node[greennode] (nine) [below=of eight] {9};
            \node[graynode] (falsethree) [above=of one] {3};
            \node[graynode] (falseseven) [above=of eight] {7};
            
            \draw[->] (two.north) -- (one.south);
            \draw[->] (one.east) -- (three.west);
            \draw[->] (four.north) -- (three.south);
            \draw[->] (five.east) -- (six.west);
            \draw[->] (eight.west) -- (seven.east);
            \draw[->] (nine.north) -- (eight.south);
            \draw[->] (falsethree.east) -- (five.west);
            \draw[->] (falseseven.west) -- (six.east);
        \end{tikzpicture}
        \caption{Full reconstruction}
        \label{fig: full reconstruction}
    \end{subfigure}
    \hfill
    \begin{subfigure}[b]{0.45\textwidth}
        \centering
        \begin{tikzpicture}[
                thick,
                bluenode/.style={circle, draw=teal, very thick, minimum size=7mm},
                magentanode/.style={circle, draw=olive, very thick, minimum size=7mm},
                greennode/.style={circle, draw=purple, very thick, minimum size=7mm},
                graynode/.style={circle, draw=darkgray, fill=gray!30, very thick, minimum size=7mm},
            ]
            \node[bluenode] (one) {1};
            \node[bluenode] (two) [below= of one] {2};
            \node[bluenode] (three) [right=of one] {3};
            \node[bluenode] (four) [below=of three] {4};
            \node[magentanode] (five) [above=of three] {5};
            \node[magentanode] (six) [right=of five] {6};
            \node[greennode] (seven) [below=of six] {7};
            \node[greennode] (eight) [right=of seven] {8};
            \node[greennode] (nine) [below=of eight] {9};
            \node[graynode] (falsethree) [above=of one] {3};
            \node[graynode] (falseseven) [above=of eight] {7};
            
            \draw[->] (two.north) -- (one.south);
            \draw[->] (one.east) -- (three.west);
            \draw[->] (four.north) -- (three.south);
            \draw[->] (five.east) -- (six.west);
            \draw[->] (eight.west) -- (seven.east);
            \draw[->] (nine.north) -- (eight.south);
            \draw[dashed, ->] (falsethree.east) -- (five.west) node[anchor=north east] {\scriptsize{p=0.5}};
            \draw[dashed, ->] (falseseven.west) -- (six.east) node[anchor=north west] {\scriptsize{p=0.5}};
        \end{tikzpicture}
        \caption{Dropout reconstruction}
        \label{fig: dropout reconstruction}
    \end{subfigure}
    \caption{Example of cluster-border neighbor reconstruction applied to the partition of \autoref{fig: division} using a 1-hop GNN. Reconstructed nodes are shown in gray. (a) \emph{Full reconstruction} method; (b) \emph{dropout reconstruction}, where the reconstructed edges are drawn with dashed lines to indicate there is a probability of restoring each edge.}
    \label{fig: reconstruction}
\end{figure}
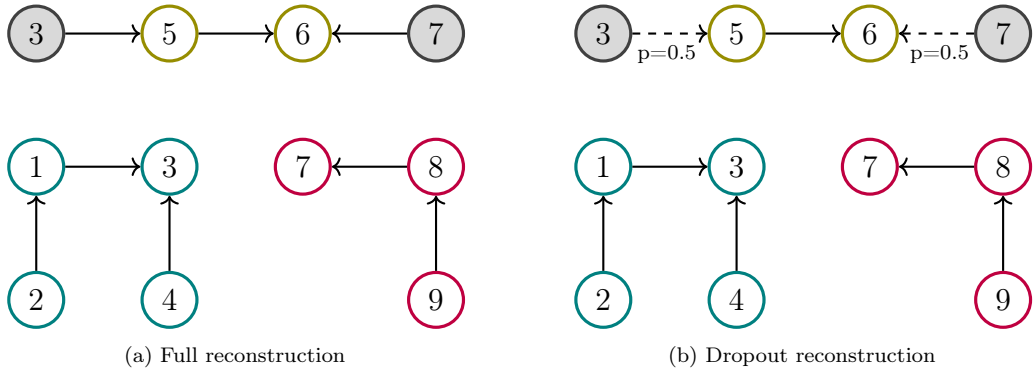

This reconstruction restores, for each cluster, all nodes and edges that were removed. In practice, this requires reconstructing as many hops as those used by the GNN. As a result, nodes at the boundaries of the cluster produce the same outputs as in the original graph (i.e.,~without edge removal). The reconstructed nodes are duplicates of the originals, increasing the overall graph size and memory footprint. Since only broken \emph{incoming} edges have to be reconstructed, in the example of \autoref{fig: full reconstruction}, where a single hop needs to be restored, only nodes~3 and~7 are added to the top cluster.

\subsubsection{Dropout reconstruction}

Although reducing computation time while preserving identical explainability outcomes is desirable, it comes at a cost: reconstructing multiple hops of neighbors can dramatically increase memory usage. To mitigate this, we propose a second strategy, termed \emph{dropout reconstruction} (\autoref{fig: dropout reconstruction}). Inspired by dropout regularization~\cite{dropout}, we reconstruct only a fraction of the broken edges, choosing to restore each edge independently with probability $p$. Thus, $p=1$ recovers all missing edges, $p=0$ recovers none, and for $0<p<1$ the expected number of reconstructed edges is $p\cdot m$, where $m$ is the total number of broken edges.

\subsection{Batch creation and explainability computation}
We create batches by selecting one non-reconstructed node per cluster so that explainability can be computed in parallel without contamination from shared neighbors. Reconstructed nodes are not included, as they are only copies used to ensure correct solutions for border nodes. The number of batches is $b=\max\{n_i\}_{i=1}^c$ where $c$ is the number of clusters and $n_i$ is the number of non-replicated nodes in cluster $i$. Hence, all node-level explanations can be computed in $b$ iterations instead of $n$, the total number of nodes in the graph (i.e.,~the classical sequential approach). 

Ideally, $b=1$; however, this would require $n$ singleton clusters and reconstruction of all broken edges, leading to an impractically large memory footprint. In general, larger clusters reduce the number of batches, but increase memory usage through reconstruction, whereas smaller clusters do the opposite. Excessive reconstruction may also increase the computation time.

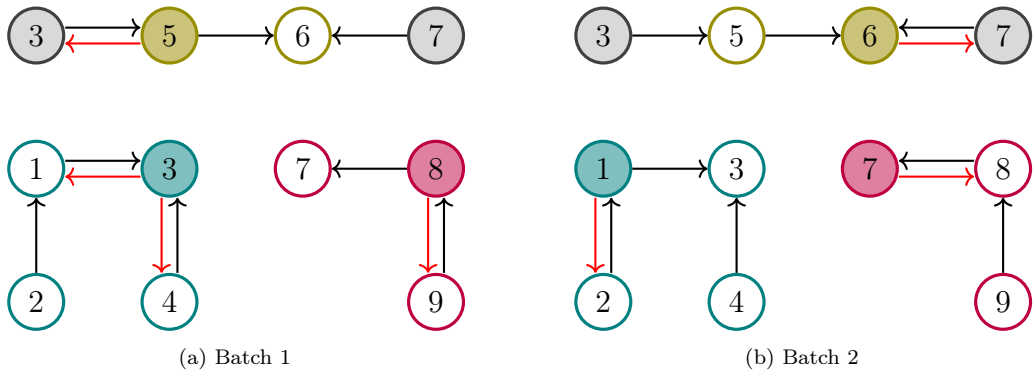
\begin{figure}
    \centering
    \begin{subfigure}[b]{0.45\textwidth}
        \centering
        \begin{tikzpicture}[
                thick,
                bluenode/.style={circle, draw=teal, very thick, minimum size=7mm},
                bluefillnode/.style={circle, draw=teal, fill=teal!50, very thick, minimum size=7mm},
                greennode/.style={circle, draw=olive, very thick, minimum size=7mm},
                greennfillode/.style={circle, draw=olive, fill=olive!50, very thick, minimum size=7mm},
                magentanode/.style={circle, draw=purple, very thick, minimum size=7mm},
                magentafillnode/.style={circle, draw=purple, fill=purple!50, very thick, minimum size=7mm},
                graynode/.style={circle, draw=darkgray, fill=gray!30, very thick, minimum size=7mm},
            ]
            \node[bluenode] (one) {1};
            \node[bluenode] (two) [below= of one] {2};
            \node[bluefillnode] (three) [right=of one] {3};
            \node[bluenode] (four) [below=of three] {4};
            \node[greennfillode] (five) [above=of three] {5};
            \node[greennode] (six) [right=of five] {6};
            \node[magentanode] (seven) [below=of six] {7};
            \node[magentafillnode] (eight) [right=of seven] {8};
            \node[magentanode] (nine) [below=of eight] {9};
            \node[graynode] (falsethree) [above=of one] {3};
            \node[graynode] (falseseven) [above=of eight] {7};
            
            \draw[->] (two.north) -- (one.south);
            \draw[->] ([yshift= 3pt] one.east) -- ([yshift= 3pt] three.west);
            \draw[->] ([xshift= 3pt] four.north) -- ([xshift= 3pt] three.south);
            \draw[->] (five.east) -- (six.west);
            \draw[->] (eight.west) -- (seven.east);
            \draw[->] ([xshift= 3pt] nine.north) -- ([xshift= 3pt] eight.south);
            \draw[->] ([yshift= 3pt] falsethree.east) -- ([yshift= 3pt] five.west);
            \draw[->] (falseseven.west) -- (six.east);
            \draw[red, ->] ([yshift= -3pt] five.west) -- ([yshift= -3pt] falsethree.east);
            \draw[red, ->] ([yshift= -3pt] three.west) -- ([yshift= -3pt] one.east);
            \draw[red, ->] ([xshift= -3pt] three.south) -- ([xshift= -3pt] four.north);
            \draw[red, ->] ([xshift= -3pt] eight.south) -- ([xshift= -3pt] nine.north);
        \end{tikzpicture}
        \caption{Batch 1}
    \end{subfigure}
    \hfill
    \begin{subfigure}[b]{0.45\textwidth}
        \centering
        \centering
        \begin{tikzpicture}[
                thick,
                bluenode/.style={circle, draw=teal, very thick, minimum size=7mm},
                bluefillnode/.style={circle, draw=teal, fill=teal!50, very thick, minimum size=7mm},
                greennode/.style={circle, draw=olive, very thick, minimum size=7mm},
                greennfillode/.style={circle, draw=olive, fill=olive!50, very thick, minimum size=7mm},
                magentanode/.style={circle, draw=purple, very thick, minimum size=7mm},
                magentafillnode/.style={circle, draw=purple, fill=purple!50, very thick, minimum size=7mm},
                graynode/.style={circle, draw=darkgray, fill=gray!30, very thick, minimum size=7mm},
            ]
            \node[bluefillnode] (one) {1};
            \node[bluenode] (two) [below= of one] {2};
            \node[bluenode] (three) [right=of one] {3};
            \node[bluenode] (four) [below=of three] {4};
            \node[greennode] (five) [above=of three] {5};
            \node[greennfillode] (six) [right=of five] {6};
            \node[magentafillnode] (seven) [below=of six] {7};
            \node[magentanode] (eight) [right=of seven] {8};
            \node[magentanode] (nine) [below=of eight] {9};
            \node[graynode] (falsethree) [above=of one] {3};
            \node[graynode] (falseseven) [above=of eight] {7};
            
            \draw[->] ([xshift= 3pt] two.north) -- ([xshift= 3pt] one.south);
            \draw[->] (one.east) -- (three.west);
            \draw[->] (four.north) -- (three.south);
            \draw[->] (five.east) -- (six.west);
            \draw[->] ([yshift= 3pt] eight.west) -- ([yshift= 3pt] seven.east);
            \draw[->] (nine.north) -- (eight.south);
            \draw[->] (falsethree.east) -- (five.west);
            \draw[->] ([yshift= 3pt] falseseven.west) -- ([yshift= 3pt] six.east);
            \draw[red, ->] ([yshift= -3pt] six.east) -- ([yshift= -3pt] falseseven.west);
            \draw[red, ->] ([xshift= -3pt] one.south) -- ([xshift= -3pt] two.north);
            \draw[red, ->] ([yshift= -3pt] seven.east) -- ([yshift= -3pt] eight.west);
        \end{tikzpicture}
        \caption{Batch 2}
    \end{subfigure}
    \caption{Computation of explainability in batches (one node per cluster), based on the example in \autoref{fig: reconstruction}. Original edges are shown in black, while red edges indicate the backward pass used for gradient computation (opposite to the forward direction). Nodes whose explainability is computed in each batch are highlighted with a colored background; all other nodes remain white.}
    \label{fig: batches}
\end{figure}

This batching process, together with the explainability computation, are illustrated in \autoref{fig: batches}, which shows the two bottom clusters of \autoref{fig: reconstruction}. In this example, the total number of iterations is~4, the maximum number of non-reconstructed nodes in any cluster. By contrast, the traditional approach requires 9~iterations. Any explainability technique can be applied within each batch without the risk of contamination from neighboring nodes. As noted earlier, explainability is computed only for non-reconstructed (i.e.,~original) nodes.

\newpage
    
\section{Theoretical complexity analysis}
\label{sec: theoretical}

Before presenting the experimental results, we provide a theoretical analysis of execution times. A natural first step is to examine the computational complexity of calculating explainability in GNN models. This complexity depends on the specific type of layer. Here, we focus on the two layers used in our experiments: GCN~\cite{gcn} and GAT~\cite{gat}. Further architectural details are given in Section~\ref{sec: results}. As discussed in~\cite{gat} and~\cite{scalable} and explained below, both layers exhibit similar computational complexity. This analysis is reproducible for any other GNN layer that one might wish to implement, allowing practitioners to derive analogous conclusions for their own architectures.

\subsection{Complexity of sequential explainability computation}

Let $G=(V,E)$ be the input graph with $n=|V|$ nodes and $m=|E|$ edges, and let $F$ denote the dimensionality of node features.  A single forward pass through a message‐passing GNN requires, at each layer, aggregating messages over all edges and then computing an update for each node. The cost per layer is therefore
\begin{equation}
\mathcal{O}(mF + nF) \sim \mathcal{O}(mF),
\end{equation}
under the common assumption that $m \ge n$ (i.e.,~the graph is not extremely sparse)~\cite{scalable}. The backward pass (backpropagation) traverses the same graph structure and also costs $\mathcal{O}(mF)$ per layer. Hence, for an $L$‐layer GNN, a single forward-background computation has complexity
\begin{equation}
\mathcal{O}\bigl(L mF + L mF\bigr) \sim \mathcal{O}(LmF).
\end{equation}
Since $L$ is typically a small constant, we write $\mathcal{O}(mF)$ for one gradient-based update over the entire graph.

As discussed in Section~\ref{sec: parallel}, to generate faithful node-level explanations, each node must be processed independently. This requires one forward pass and one backward pass per node. Consequently, the total complexity of computing explainability sequentially becomes
\begin{equation}
\mathcal{O}(nmF).
\end{equation}

\subsection{Complexity of parallel cluster-wise explainability computation}
Our method first partitions the node set $V$ into $c$ disjoint clusters $V_1,\dots,V_c$, each of size approximately\footnote{All clusters are assumed to be of equal size, which is fairly consistent with our experiments using METIS.} $n/c$. We then build an expanded disconnected graph $G'=(V',E')$ by replicating edges so that node‐level explanations can be computed independently within each cluster. Let 
\begin{equation}
m = |E|,\quad
r = |E'|-|E|
\end{equation}
so that the total number of edges in the reconstructed graph $G'$ is $m' = m + r$, that is, the number of edges in the original graph plus the number of reconstructed edges. Although $|V'|\geq|V|$, the number of edges dominates over the number of nodes and, therefore, does not materially affect the computational complexity.

Because the $c$ clusters can be processed in parallel, the number of forward-backward passes is given by the size of the largest cluster, approximately $n/c$. Therefore, the complexity of computing explainability using the proposed method is
\begin{equation}
\mathcal{O}\left( \frac{n}{c} m' F \right) = \mathcal{O}\left( \frac{n}{c} (m + r) F \right)
\label{eq:complexity2}
\end{equation}

Note that we neglect the cost of graph partitioning and reconstruction. These operations should be included in a complete cost model, but they represent constant or lower‐order terms relative to $\mathcal{O}\left(\frac{n}{c}m'F\right)$.

\subsubsection{When is parallelization beneficial?}
\label{sec: payoff}
To obtain a speedup over the sequential baseline (i.e.,~$\mathcal{O}(n\,mF)$), the cost of our parallel strategy must satisfy
\begin{equation}
\frac{n}{c}(m + r)F \;<\; n m F \iff m + r \;<\; c m \iff c > \frac{r}{m} + 1.
    \label{eq:sppedup}
\end{equation}
In other words, the number of clusters must exceed $r/m + 1$, where $r/m$ is the \emph{edge‐replication ratio}. This ratio quantifies how many reconstructed edges are added relative to the number of original graph edges. In practice, it is usually greater than one because the same border edges are reconstructed across multiple clusters. Moreover, both $c$ and $r$ tend to increase together. Nevertheless, as empirically shown in Section~\ref{sec: results}, the inequality is typically satisfied as $c$ grows.

In practice, the advantage is usually even greater. Forward and backward passes in modern deep-learning frameworks are implemented in optimized low-level languages (e.g.,~C\texttt{++}), whereas the loop over $n$ sequential explainability runs is driven by higher-level code (e.g.,~Python). Since inference is heavily optimized~\cite{torch2} and low-level languages are generally much faster than their higher-level counterparts~\cite{language_comparison}, reducing the number of sequential operations yields additional performance gains beyond the asymptotic bound. This explains why the time reductions observed in Section~\ref{sec: results} are larger what the condition $c > r/m + 1$ would predict.

\section{Experimental results and discussion}
\label{sec: results}

This section presents a representative subset of the experimental results and discusses their implications (see \ref{sec: complete results} for the full results). We use the citation network datasets Cora, CiteSeer, and PubMed~\cite{cora}, which are commonly employed in node-classification tasks and explainability benchmarks~\cite{evaluatinggnnxai}. For the models, we test a single architecture with two widely used GNN layers: GCN~\cite{gcn} and GATv2~\cite{gat}. The architecture consists of two GNN layers with ReLU activations and Dropout~\cite{dropout} between them. This design reflects typical classification architectures while keeping the number of layers small.

All experiments were conducted on a machine running Ubuntu 20.04.6 LTS, equipped with an NVIDIA Tesla~V100 GPU with \qty{16}{GB} VRAM, an Intel\textsuperscript{\textregistered} Xeon\textsuperscript{\textregistered} Gold 6238R CPU (8~cores at 2.20GHz), and \qty{15}{GB} RAM. The code~\footnote{The source code is available at \url{https://github.com/EricssonResearch/parallel-xai-gnn}} was implemented in Python 3.12 using PyTorch~\cite{torch} 2.6.0. Each experiment was repeated three times, and execution times are averaged to provide more reliable estimates.

\autoref{tab: parallelization results} summarizes the performance gains of the proposed parallelization strategy for the Saliency Map explainability method applied to the GAT~model using 16~clusters. This configuration was chosen because it does not encounter memory limitations on the largest dataset, PubMed. Execution times are substantially reduced across all datasets, with improvements exceeding 80\% for Cora and CiteSeer and remaining above 50\% for PubMed. These results demonstrate the effectiveness of our approach in accelerating explainability computations, and even larger gains are expected for more clusters. For the remaining examples in this section, we use GNNExplainer, the most widely adopted GNN explainability method. Unfortunately, it cannot be run on PubMed with our current hardware setup.

\begin{table}[t]
  \caption{Execution time comparison between the original sequential approach and the proposed parallelized method using full reconstruction. Results correspond to the Saliency Map explainability method applied to the GAT model with 16 clusters across all datasets. The gain indicates the relative reduction in execution time, with 50\%  gain corresponding to a halving of the original runtime.
  }
  \label{tab: parallelization results}
  \centering
  \scalebox{0.8}{
  \begin{tabular}{cccc}
    \toprule
    Dataset & Original time (s) & Parallelized method time (s) & Gain (\%) \\
    \midrule
    Cora & 28.40 & 4.30 &  84.86 \\
    CiteSeer & 35.65 & 3.57 & 89.99 \\
    PubMed & 401.21 & 182.07 & 54.62 \\
    \bottomrule
  \end{tabular}
  }
\end{table}

The number of clusters is a key parameter in the proposed method, as it directly affects overall execution time. Increasing the number of clusters allows for greater parallelization, but also increases the reconstruction overhead at cluster borders. To illustrate this trade-off, \autoref{tab: full reconstruction gnnexplainer cora gat} reports the following metrics obtained for different numbers of clusters:

\begin{itemize}
    \item Execution time: Average over three runs to compute the explainability for all nodes in the dataset.
    \item $\Delta$nodes~(\%): Percentage of additional nodes added in the full reconstruction relative to the number of nodes in the original graph. A value of~100\% indicates that the number of nodes has doubled, while~0\% means that no nodes were reconstructed (i.e.,~no partitioning). 
    \item $\Delta$edges~(\%): Analogous to $\Delta$nodes, this quantity represents the percentage increase in the number of edges.
\end{itemize}

\begin{table}[t]
  \caption{Full reconstruction using GNNExplainer on the Cora dataset with the GAT model.}
  \label{tab: full reconstruction gnnexplainer cora gat}
  \centering
  \scalebox{0.8}{
  \begin{tabular}{cccccc}
    \toprule
    Number of clusters & Execution time (s) & $\Delta$nodes (\%) & $\Delta$edges (\%) \\
    \midrule
    1 & 3069.01 & 0.00 & 0.00 \\
    8 & 573.87 & 399.08 & 428.95 \\
    16 & 394.05 & 720.35 & 774.97 \\
    32 & 251.06 & 1084.86 & 1149.77 \\
    64 & 165.26 & 1667.25 & 1736.23 \\
    128 & 127.77 & 2774.52 & 2860.34 \\
    \bottomrule
  \end{tabular}
  }
\end{table}

\begin{figure}[t]
  \centering
  \begin{subfigure}{0.49\linewidth}
      \centering
      \includegraphics[width=\linewidth, height=0.8\linewidth]{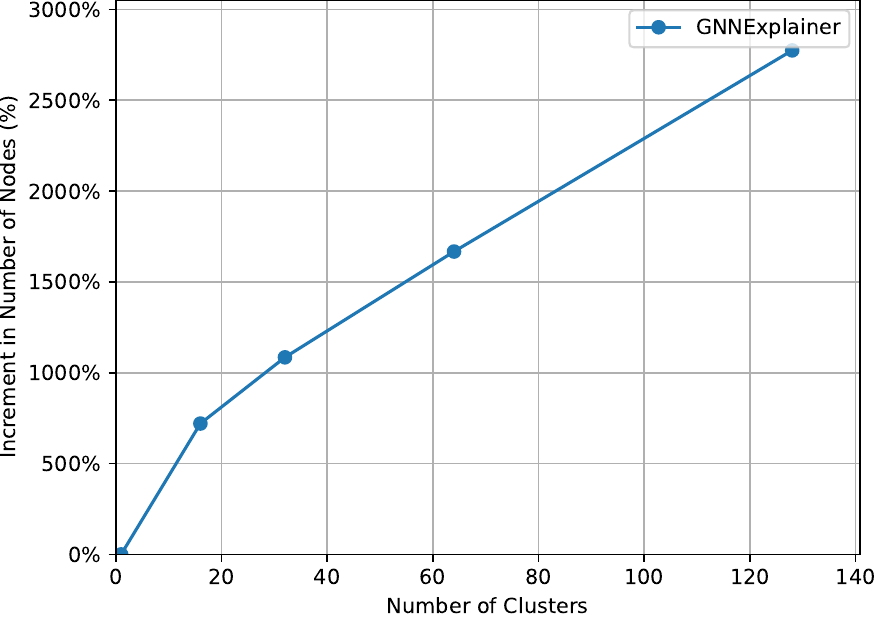}
      \caption{Increase in number of nodes.}
  \end{subfigure}
  \begin{subfigure}{0.49\linewidth}
    \centering
    \includegraphics[width=\linewidth, height=0.8\linewidth]{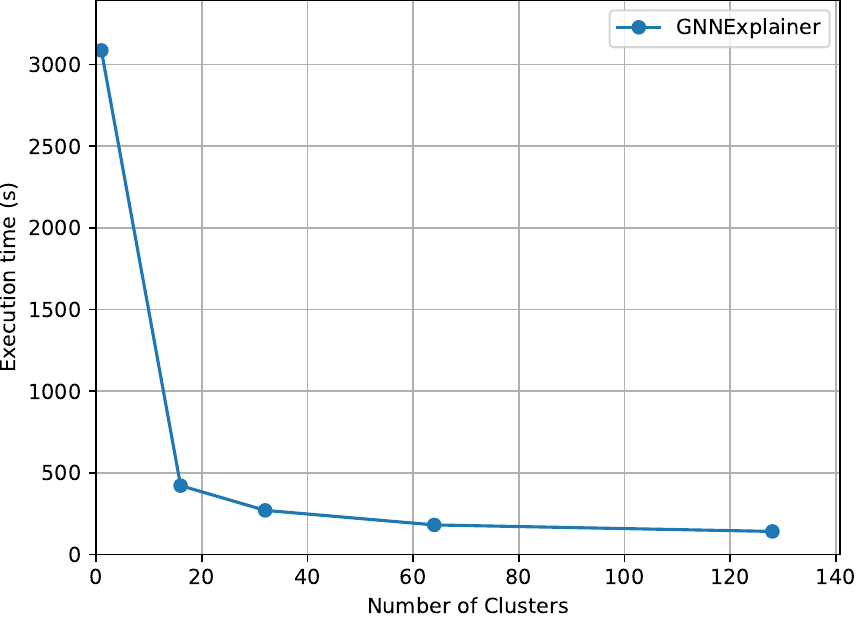}
    \caption{Decrease in execution time.}
  \end{subfigure}
  \caption{Evolution of (a) the number of nodes and (b) the execution time depending on the number of clusters for \emph{full reconstruction} with GNNExplainer on the Cora dataset and the GAT~model.}
  \label{fig: graphs full reconstruction}
\end{figure}

As the number of clusters grows, execution times decrease consistently in \autoref{tab: full reconstruction gnnexplainer cora gat}, but at the cost of higher memory usage due to the additional reconstructed nodes and edges. Despite the substantial growth in graph size, the overall execution time continues to decrease, showing that the benefits of increased parallelization outweigh the reconstruction overhead. These trends are further illustrated in \autoref{fig: graphs full reconstruction} and are consistent across all experiments (see \ref{sec: complete results}). These results suggest diminishing returns in execution time reduction beyond a certain number of clusters, as the benefits of additional parallelization progressively balance the increasing reconstruction cost.

The optimal number of clusters and the associated memory footprint ultimately depend on the application. The number of reconstructed nodes is influenced not only by the chosen partitioning technique, but also by the intrinsic topology of the graph. In densely connected graphs, more edges are cut during partitioning, which increases reconstruction and therefore memory usage. Although identifying an optimal cluster count lies outside the scope of this paper, a practical guideline is to use the largest number of clusters that can be supported by the available memory resources.

\begin{table}[p]
\caption{Dropout reconstruction with GNNExplainer on the Cora dataset with the GAT model.}
\label{tab: dropout reconstruction gnnexplainer cora gat}
\centering
\scalebox{0.7}{
    \begin{tabular}{ccccccccccc}
        \toprule
        \thead{Number \\ of clusters} &
        \thead{Dropout \\ rate} &
        \thead{Execution \\ time (s)} & 
        \thead{$\Delta$nodes \\ (\%)} & 
        \thead{$\Delta$edges \\ (\%)} &  
        \multicolumn{3}{c}{\thead{Affected \\ explanations (\%)}} &  
        \multicolumn{3}{c}{\thead{Affected \\ nodes (\%)}}  \\
        \cmidrule(lr){6-8}
        \cmidrule(lr){9-11}
        & & & & & 0.2 & 0.5 & 0.7 & 0.2 & 0.5 & 0.7 \\
        \midrule

        \multirow{1}{*}{1} & 0.00 & 3069.01 &    0.00 &    0.00 &  0.00 &  0.00 &  0.00 &  0.00 &  0.00 &  0.00 \\
        \midrule

        \multirow{5}{*}{8} 
            & 0.00 & 573.87  & 399.08 & 428.95 & 71.64 & 28.25 & 0.07 & 6.96  & 1.29 & 0.00 \\
            & 0.20 & 506.75  & 317.61 & 276.68 & 81.68 & 36.93 & 11.23 & 14.34 & 2.11 & 0.55 \\
            & 0.50 & 440.40  & 200.85 & 102.54 & 86.74 & 46.12 & 23.93 & 27.32 & 3.78 & 1.60 \\
            & 0.70 & 413.24  & 120.72 &  32.97 & 88.18 & 47.19 & 26.66 & 30.05 & 4.10 & 1.81 \\
            & 1.00 & 387.25  &   0.00 & -10.06 & 88.85 & 51.18 & 31.24 & 35.67 & 4.89 & 2.33 \\
        \midrule

        \multirow{5}{*}{16}
          & 0.00 & 394.05  & 720.35 & 774.97 & 74.70 & 30.69 & 0.04 & 7.30  & 1.39 & 0.00 \\
          & 0.20 & 325.46  & 574.56 & 498.24 & 87.67 & 41.40 & 16.99 & 17.71 & 2.58 & 0.81 \\
          & 0.50 & 247.46  & 365.14 & 186.83 & 89.33 & 52.03 & 31.35 & 36.11 & 4.84 & 2.35 \\
          & 0.70 & 216.90  & 215.21 &  65.31 & 89.33 & 52.47 & 33.53 & 39.40 & 5.46 & 2.82 \\
          & 1.00 & 193.93  &   0.00 & -13.93 & 90.32 & 56.72 & 41.03 & 51.02 & 6.66 & 3.67 \\
        \midrule

        \multirow{5}{*}{32} 
          & 0.00 & 251.06  & 1084.86 & 1149.77 & 74.37 & 30.61 & 0.00 & 7.33  & 1.44 & 0.00 \\
          & 0.20 & 204.49  & 869.76  & 743.69  & 87.37 & 44.76 & 19.90 & 22.73 & 3.29 & 1.32 \\
          & 0.50 & 143.74  & 542.10  & 288.97  & 90.73 & 55.76 & 38.29 & 43.01 & 5.80 & 3.12 \\
          & 0.70 & 115.40  & 321.82  &  88.92  & 90.92 & 58.83 & 44.31 & 55.95 & 7.12 & 4.15 \\
          & 1.00 &  97.30  &   0.00  & -18.57  & 91.51 & 61.67 & 49.63 & 63.74 & 8.00 & 4.88 \\
        \midrule

        \multirow{5}{*}{64}
          & 0.00 & 165.26  & 1667.25 & 1736.23 & 77.22 & 32.94 & 0.04 & 7.69  & 1.57 & 0.00 \\
          & 0.20 & 131.11  & 1333.27 & 1101.27 & 91.10 & 49.59 & 28.21 & 27.83 & 3.81 & 1.71 \\
          & 0.50 &  87.26  &  838.88 &  433.10 & 92.32 & 58.53 & 44.83 & 49.66 & 6.43 & 3.60 \\
          & 0.70 &  64.72  &  509.60 &  150.06 & 91.95 & 60.49 & 48.45 & 59.80 & 7.56 & 4.64 \\
          & 1.00 &  48.39  &    0.00 &  -24.40 & 91.47 & 65.81 & 56.79 & 73.18 & 9.13 & 5.94 \\
        \midrule

        \multirow{5}{*}{128}
          & 0.00 & 127.77  & 2774.52 & 2860.34 & 77.07 & 32.79 & 0.04 & 7.61  & 1.56 & 0.00 \\
          & 0.20 &  96.86  & 2220.86 & 1832.40 & 94.31 & 58.01 & 38.88 & 33.58 & 4.80 & 2.42 \\
          & 0.50 &  60.03  & 1393.13 &  704.13 & 94.61 & 68.80 & 59.19 & 61.60 & 8.79 & 5.56 \\
          & 0.70 &  40.67  &  825.96 &  222.13 & 93.39 & 71.75 & 65.36 & 77.06 &10.71 & 7.27 \\
          & 1.00 &  25.22  &    0.00 &  -50.42 & 92.28 & 75.89 & 71.16 & 87.57 &12.65 & 9.02 \\
        \bottomrule
    \end{tabular}
}
\end{table}

Motivated by the memory overhead introduced by \emph{full reconstruction}, we propose a \emph{dropout-based reconstruction} strategy that selectively reconstructs only a fraction of border nodes according to a given probability. This reduces memory requirements at the expense of a decrease in explainability fidelity. To empirically characterize this trade-off between computational efficiency and explanation quality, we vary the dropout rate from 0 to 1 and evaluate its effect using two metrics:

\begin{itemize}
    \item Affected nodes: Nodes whose explainability scores under \emph{dropout reconstruction} differ from those obtained with \emph{full reconstruction} by more than a specified threshold. Scores are normalized to the $[0,1]$ interval for each target node, where~1 corresponds to the most immportant node in the explanation and 0 to the least important. To reflect different tolerance levels, we consider thresholds of $0.2$, $0.5$, and $0.7$.
    \item Affected explanations: Percentage of node-level explanations that contain at least one \emph{affected node}. This metric captures cases where even a small number of perturbed nodes may undermine the reliability of the entire explanation.
\end{itemize}

\begin{figure}[t]
  \centering
  \begin{subfigure}{0.49\linewidth}
      \centering
      \includegraphics[width=\linewidth, height=0.8\linewidth]{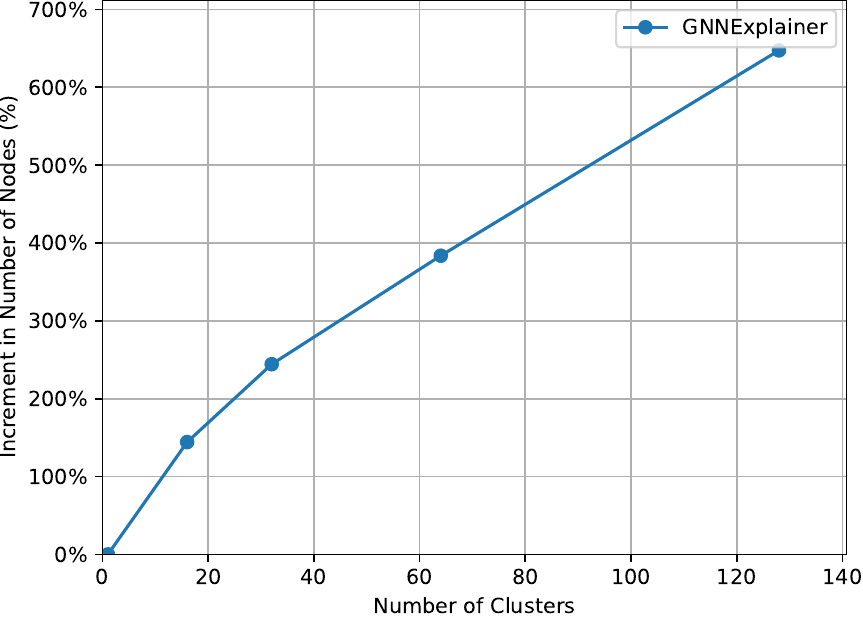}
      \caption{Increase in number of nodes.}
  \end{subfigure}
  \begin{subfigure}{0.49\linewidth}
    \centering
    \includegraphics[width=\linewidth, height=0.8\linewidth]{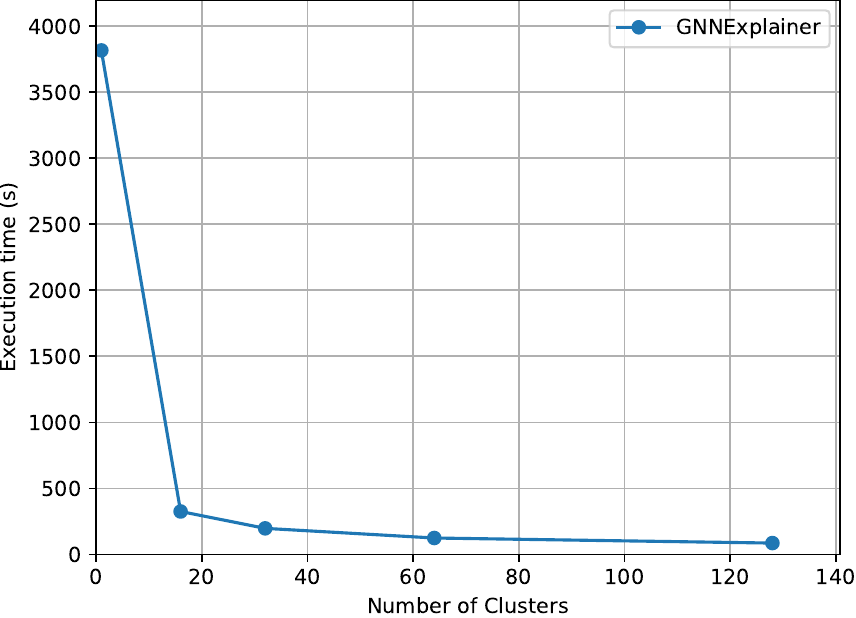}
    \caption{Decrease in execution time.}
  \end{subfigure}
  \caption{Evolution of (a)~the number of nodes and (b)~the computational time depending on the number of clusters for \emph{full reconstruction} with GNNExplainer on the CiteSeer dataset and the GAT model.}
  \label{fig: graphs full reconstruction CiteSeer}
\end{figure}

\autoref{tab: dropout reconstruction gnnexplainer cora gat} shows that, for a fixed number of clusters, increasing the dropout rate consistently reduces both execution time and reconstruction overhead, since fewer nodes and edges are reinstated at cluster borders. However, higher dropout rates also increase the risk of discarding nodes that are important for the explanation, leading to deviations from the results obtained on the original graph. Unfortunately, identifying these key nodes in advance would require computing the explainability scores sequentially, the very computation that the parallelization aims to avoid.

While the previous analyses focused on the Cora dataset, the memory overhead is lower for the other datasets. As depicted in \autoref{fig: graphs full reconstruction CiteSeer}, reconstruction still increases memory usage in CiteSeer, but the effect is less pronounced than in Cora. These dataset-specific differences highlight the importance of tuning hyperparameters, such as the number of clusters or the dropout rate, for each scenario. The full set of results is provided in \ref{sec: complete results}.

\section{Conclusion and future work}
\label{sec: conclusion}

This paper examines the benefits of parallelization for computing node-level explainability in GNNs. The main contributions can be summarized as follows:

\begin{itemize}
    \item Identification of the batch-execution bottleneck. We show that many explainability techniques cannot be executed in batches on GNNs. In particular, we highlight the ``gradient-mixing'' problem, which prevents gradient-based methods, including widely used techniques such as GNNExplainer, from producing prediction-specific explanations when run in parallel.
    
    \item A general parallelization framework. We introduce a technique that enables explainability to be computed in parallel by constructing batches of independent nodes. The method is agnostic to the underlying explainability algorithm (e.g.,~GNNExplainer, Saliency Maps) and yields exactly the same explanations as the sequential baseline. Both theoretical analysis and empirical evaluation show that this framework delivers substantial reductions in execution time, making node-level explainability feasible in production-scale environments.
    
    \item A memory–efficiency mechanism. Because graph partitioning requires node and edge reconstruction at cluster boundaries, we also propose \emph{dropout reconstruction}, which probabilistically restores only a fraction of the broken edges. This allows users to balance memory usage against explanation accuracy.
\end{itemize}

While METIS was chosen as the partitioning tool due to its widespread adoption in GNN research, the resulting memory overhead suggests that further studies are needed to identify alternative partitioning strategies that better trade off reconstruction cost and speed. Besides, approaches that eliminate the clustering stage altogether may remove the need for reconstruction entirely. They represent a promising direction for future research toward even more efficient GNN explainability.



\section*{CRediT authorship contribution statement}
\textbf{Oscar Llorente:} Writing – original draft, Software, Methodology, Conceptualization; \textbf{Jaime Boal:} Writing – review
\& editing, Conceptualization, Supervision; \textbf{Eugenio F. Sánchez-Úbeda:} Writing – review \& editing, Conceptualization, Visualization, Supervision; \textbf{Antonio Diaz-Cano:} Formal analysis, Visualization; \textbf{Miguel Familiar:} Writing – review
\& editing

\section*{Declaration of AI-assisted technologies in the writing process}
The authors used tools such as LanguageTool, Writefull, and ChatGPT to correct spelling and enhance clarity, grammar, and sentence structure during the manuscript preparation. Subsequently, the authors thoroughly reviewed and edited the content as required, assuming full responsibility for the publication's content.

\section*{Code availability}

The source code is available at \url{https://github.com/EricssonResearch/parallel-xai-gnn}.

\section*{Declaration of competing interest}

The authors declare that they have no known competing financial
interests or personal relationships that could have appeared to influence
the work reported in this paper.

\bibliographystyle{elsarticle-num} 
{\small\bibliography{references}}

\appendix

\section{Complete results}
\label{sec: complete results}

This appendix ppresents the results obtained for two models (GCN~\cite{gcn} and GAT~\cite{gat}) using four explainability techniques (DeconvNet~\cite{zeiler}, GNNExplainer~\cite{gnnexplainer}, Guided Backprop~\cite{SpringenbergDBR14}, and Saliency Maps~\cite{simonyanDeepConvolutionalNetworks2014a}) on three citation network datasets: Cora, CiteSeer, and PubMed~\cite{cora}.

\autoref{tab: summary_refs_tables} serves as an index to all detailed result tables corresponding to each combination of explainability technique, dataset, and model architecture. For every combination, we evaluate multiple clustering configurations (ranging from 1 to 128 clusters) and vary the dropout rate between 0 and 1. Reported metrics include execution time, increases in nodes and edges, and the percentage of affected explanations and nodes.

Note that for the PubMed dataset, which is significantly larger than Cora and CiteSeer, experiments are limited to 16 clusters due to the GPU memory required to store the reconstructed dataset. Running larger cluster configurations on CPU would prevent a fair comparison with the GPU-based experiments on the smaller datasets, since CPU execution times would be disproportionately higher. Similarly, GNNExplainer is not run on PubMed because its memory requirements exceed the available GPU capacity.

\begin{table}[hb]
    \caption{Summary of table references for each explainability technique evaluated on the Cora, CiteSeer, and PubMed datasets under~GCN and~GAT model architectures.}
    \label{tab: summary_refs_tables}
    \centering
    \scalebox{0.75}{
\begin{tabular}{lcccccc}
    \toprule
    \textbf{Explainability}
    & \multicolumn{2}{c}{\textbf{Cora}} 
    & \multicolumn{2}{c}{\textbf{CiteSeer}} 
    & \multicolumn{2}{c}{\textbf{PubMed}} \\
    \cmidrule(lr){2-3} \cmidrule(lr){4-5} \cmidrule(lr){6-7}
    \textbf{technique} & GCN & GAT & GCN & GAT & GCN & GAT \\
    \midrule
DeconvNet & \autoref{tab: Deconvnet-Cora-GCN} & \autoref{tab: Deconvnet-Cora-GAT} & \autoref{tab: Deconvnet-CiteSeer-GCN} &  \autoref{tab: Deconvnet-CiteSeer-GAT} & \autoref{tab: Deconvnet-PubMed-GCN} & \autoref{tab: Deconvnet-PubMed-GAT} \\
GNNExplainer & \autoref{tab: GNNExplainer-Cora-GCN} & \autoref{tab: GNNExplainer-Cora-GAT} & \autoref{tab: GNNExplainer-CiteSeer-GCN} &  \autoref{tab: GNNExplainer-CiteSeer-GAT} & -- &  -- \\
Guided Backprop & \autoref{tab: GuidedBackprop-Cora-GCN} & \autoref{tab: GuidedBackprop-Cora-GAT} & \autoref{tab: GuidedBackprop-CiteSeer-GCN} & \autoref{tab: GuidedBackprop-CiteSeer-GAT} & \autoref{tab: GuidedBackprop-PubMed-GCN} & \autoref{tab: GuidedBackprop-PubMed-GAT} \\
Saliency Map & \autoref{tab: SaliencyMap-Cora-GCN} & \autoref{tab: SaliencyMap-Cora-GAT} & \autoref{tab: SaliencyMap-CiteSeer-GCN} & \autoref{tab: SaliencyMap-CiteSeer-GAT} & \autoref{tab: SaliencyMap-PubMed-GCN} & \autoref{tab: SaliencyMap-PubMed-GAT} \\
\bottomrule
\end{tabular}
}
\end{table}


\begin{table}

  \caption{Dropout reconstruction with DeconvNet on Cora dataset using the GCN model.}
  \label{tab: Deconvnet-Cora-GCN}
  \centering
  \scalebox{0.7}{
      \begin{tabular}{ccccccccccc}
        \toprule
        \thead{Number \\ of clusters} &
        \thead{Dropout \\ rate} &
        \thead{Execution \\ time (s)} & 
        \thead{$\Delta$nodes \\ (\%)} & 
        \thead{$\Delta$edges \\ (\%)} &  
        \multicolumn{3}{c}{\thead{Affected \\ explanations (\%)}} &  
        \multicolumn{3}{c}{\thead{Affected \\ nodes (\%)}}  \\
        \cmidrule(lr){6-8}
        \cmidrule(lr){9-11}
        & & & & & 0.2 & 0.5 & 0.7 & 0.2 & 0.5 & 0.7 \\
        \midrule
        
        \multirow{1}{*}{1} 
            & 0.00 & 17.01 & 0.00 & 0.00 & 0.00 & 0.00 & 0.00 & 0.00 & 0.00 & 0.00 \\
        \midrule

        \multirow{5}{*}{16} 
            & 0.00 & 3.07 & 720.35 & 774.97 & 0.00 & 0.00 & 0.00 & 0.00 & 0.00 & 0.00 \\
            & 0.20 & 2.61 & 576.00 & 498.22 & 10.78 & 1.14 & 0.37 & 0.46 & 0.03 & 0.01 \\
            & 0.50 & 1.94 & 361.63 & 182.91 & 22.01 & 2.73 & 0.41 & 1.20 & 0.08 & 0.01 \\
            & 0.70 & 1.58 & 218.28 & 60.21 & 27.66 & 4.21 & 1.33 & 1.64 & 0.12 & 0.04 \\
            & 1.00 & 1.34 & 0.00 & -13.93 & 31.09 & 5.83 & 1.44 & 2.04 & 0.17 & 0.04 \\
\bottomrule
      \end{tabular}
    }
\end{table}

\begin{table}
  \caption{Dropout reconstruction with GNNExplainer on the Cora dataset using the GCN model.}
  \label{tab: GNNExplainer-Cora-GCN}
  \centering
  \scalebox{0.7}{
      \begin{tabular}{ccccccccccc}
        \toprule
        \thead{Number \\ of clusters} &
        \thead{Dropout \\ rate} &
        \thead{Execution \\ time (s)} & 
        \thead{$\Delta$nodes \\ (\%)} & 
        \thead{$\Delta$edges \\ (\%)} &  
        \multicolumn{3}{c}{\thead{Affected \\ explanations ( \%)}} &  
        \multicolumn{3}{c}{\thead{Affected \\ nodes (\%)}}  \\
        \cmidrule(lr){6-8}
        \cmidrule(lr){9-11}
        & & & & & 0.2 & 0.5 & 0.7 & 0.2 & 0.5 & 0.7 \\
        \midrule
        \multirow{1}{*}{1} 
             & 0.00 & 1903.56 & 0.00 & 0.00 & 0.00 & 0.00 & 0.00 & 0.00 & 0.00 & 0.00 \\
        \midrule
        
        \multirow{5}{*}{16} 
            & 0.00 & 299.03 & 720.35 & 774.97 & 83.12 & 49.37 & 0.00 & 12.22 & 4.60 & 0.00 \\
            & 0.20 & 242.98 & 579.39 & 504.32 & 91.03 & 57.75 & 19.09 & 22.33 & 6.07 & 1.30 \\
            & 0.50 & 170.69 & 359.38 & 188.73 & 92.32 & 60.89 & 31.31 & 39.49 & 8.74 & 3.55 \\
            & 0.70 & 141.48 & 214.40 & 66.62 & 92.54 & 61.34 & 34.64 & 47.10 & 9.66 & 4.42 \\
            & 1.00 & 120.62 & 0.00 & -13.93 & 92.36 & 60.86 & 38.74 & 54.10 & 10.80 & 5.44 \\
\bottomrule
      \end{tabular}
    }
\end{table}

\begin{table}
  \caption{Dropout reconstruction with Guided Backprop on the Cora dataset using the GCN model.}
  \label{tab: GuidedBackprop-Cora-GCN}
  \centering
  \scalebox{0.7}{
      \begin{tabular}{ccccccccccc}
        \toprule
        \thead{Number \\ of clusters} &
        \thead{Dropout \\ rate} &
        \thead{Execution \\ time (s)} & 
        \thead{$\Delta$nodes \\ (\%)} & 
        \thead{$\Delta$edges \\ (\%)} &  
        \multicolumn{3}{c}{\thead{Affected \\ explanations (\%)}} &  
        \multicolumn{3}{c}{\thead{Affected \\ nodes (\%)}}  \\
        \cmidrule(lr){6-8}
        \cmidrule(lr){9-11}
        & & & & & 0.2 & 0.5 & 0.7 & 0.2 & 0.5 & 0.7 \\
        \midrule
        \multirow{1}{*}{1} 
          & 0.00 & 16.82 & 0.00 & 0.00 & 0.00 & 0.00 & 0.00 & 0.00 & 0.00 & 0.00 \\
        \midrule

        \multirow{5}{*}{16} 
          & 0.00 & 3.09 & 720.35 & 774.97 & 0.00 & 0.00 & 0.00 & 0.00 & 0.00 & 0.00 \\
          & 0.20 & 2.61 & 573.38 & 487.86 & 12.37 & 0.96 & 0.30 & 0.54 & 0.03 & 0.01 \\
          & 0.50 & 1.93 & 361.26 & 191.27 & 21.57 & 2.95 & 0.63 & 1.17 & 0.08 & 0.02 \\
          & 0.70 & 1.59 & 215.58 & 59.97 & 27.62 & 3.99 & 0.96 & 1.65 & 0.11 & 0.03 \\
          & 1.00 & 1.31 & 0.00 & -13.93 & 31.06 & 5.83 & 1.44 & 2.04 & 0.17 & 0.04 \\
\bottomrule
      \end{tabular}
    }
\end{table}

\begin{table}
  \caption{Dropout reconstruction with Saliency Map on Cora dataset using the GCN model.}
  \label{tab: SaliencyMap-Cora-GCN}
  \centering
  \scalebox{0.7}{
      \begin{tabular}{ccccccccccc}
        \toprule
        \thead{Number \\ of clusters} &
        \thead{Dropout \\ rate} &
        \thead{Execution \\ time (s)} & 
        \thead{$\Delta$nodes \\ (\%)} & 
        \thead{$\Delta$edges \\ (\%)} &  
        \multicolumn{3}{c}{\thead{Affected \\ explanations (\%)}} &  
        \multicolumn{3}{c}{\thead{Affected \\ nodes (\%)}}  \\
        \cmidrule(lr){6-8}
        \cmidrule(lr){9-11}
        & & & & & 0.2 & 0.5 & 0.7 & 0.2 & 0.5 & 0.7 \\
        \midrule
        \multirow{1}{*}{1} 
          & 0.00 & 16.96 & 0.00 & 0.00 & 0.00 & 0.00 & 0.00 & 0.00 & 0.00 & 0.00 \\
        \midrule

        \multirow{5}{*}{16} 
          & 0.00 & 2.89 & 720.35 & 774.97 & 0.00 & 0.00 & 0.00 & 0.00 & 0.00 & 0.00 \\
          & 0.20 & 2.49 & 576.00 & 498.22 & 10.78 & 1.14 & 0.41 & 0.46 & 0.03 & 0.01 \\
          & 0.50 & 1.85 & 361.63 & 182.91 & 22.23 & 3.06 & 0.44 & 1.22 & 0.09 & 0.01 \\
          & 0.70 & 1.56 & 218.28 & 60.21 & 27.92 & 4.36 & 1.40 & 1.66 & 0.13 & 0.04 \\
          & 1.00 & 1.33 & 0.00 & -13.93 & 31.28 & 6.13 & 1.51 & 2.05 & 0.18 & 0.04 \\
\bottomrule
      \end{tabular}
    }
\end{table}

\begin{table}
  \caption{Dropout reconstruction with DeconvNet on the Cora dataset using the GAT model.}
  \label{tab: Deconvnet-Cora-GAT}
  \centering
  \scalebox{0.7}{
      \begin{tabular}{ccccccccccc}
        \toprule
        \thead{Number \\ of clusters} &
        \thead{Dropout \\ rate} &
        \thead{Execution \\ time (s)} & 
        \thead{$\Delta$nodes \\ (\%)} & 
        \thead{$\Delta$edges \\ (\%)} &  
        \multicolumn{3}{c}{\thead{Affected \\ explanations (\%)}} &  
        \multicolumn{3}{c}{\thead{Affected \\ nodes (\%)}}  \\
        \cmidrule(lr){6-8}
        \cmidrule(lr){9-11}
        & & & & & 0.2 & 0.5 & 0.7 & 0.2 & 0.5 & 0.7 \\
        \midrule
        \multirow{1}{*}{1} 
          & 0.00 & 27.32 & 0.00 & 0.00 & 0.00 & 0.00 & 0.00 & 0.00 & 0.00 & 0.00 \\
        \midrule

        \multirow{5}{*}{16} 
          & 0.00 & 4.47 & 720.35 & 774.97 & 0.00 & 0.00 & 0.00 & 0.00 & 0.00 & 0.00 \\
          & 0.20 & 3.80 & 579.06 & 512.77 & 9.45 & 6.79 & 6.09 & 0.82 & 0.40 & 0.29 \\
          & 0.50 & 2.79 & 364.62 & 191.82 & 23.23 & 16.65 & 14.55 & 2.12 & 0.96 & 0.67 \\
          & 0.70 & 2.44 & 215.03 & 65.95 & 30.50 & 22.90 & 20.46 & 2.86 & 1.40 & 1.02 \\
          & 1.00 & 2.06 & 0.00 & -13.93 & 36.19 & 27.03 & 24.34 & 3.47 & 1.65 & 1.22 \\
\bottomrule
      \end{tabular}
    }
\end{table}

\begin{table}
  \caption{Dropout reconstruction with GNNExplainer on the Cora dataset using the GAT model.}
  \label{tab: GNNExplainer-Cora-GAT}
  \centering
  \scalebox{0.7}{
      \begin{tabular}{ccccccccccc}
        \toprule
        \thead{Number \\ of clusters} &
        \thead{Dropout \\ rate} &
        \thead{Execution \\ time (s)} & 
        \thead{$\Delta$nodes \\ (\%)} & 
        \thead{$\Delta$edges \\ (\%)} &  
        \multicolumn{3}{c}{\thead{Affected \\ explanations (\%)}} &  
        \multicolumn{3}{c}{\thead{Affected \\ nodes (\%)}}  \\
        \cmidrule(lr){6-8}
        \cmidrule(lr){9-11}
        & & & & & 0.2 & 0.5 & 0.7 & 0.2 & 0.5 & 0.7 \\
        \midrule
        \multirow{1}{*}{1} 
          & 0.00 & 3086.39 & 0.00 & 0.00 & 0.00 & 0.00 & 0.00 & 0.00 & 0.00 & 0.00 \\
        \midrule

        \multirow{5}{*}{16} 
          & 0.00 & 421.13 & 720.35 & 774.97 & 73.86 & 31.17 & 0.11 & 7.37 & 1.40 & 0.00 \\
          & 0.20 & 344.39 & 575.74 & 495.36 & 87.63 & 43.06 & 18.32 & 18.80 & 2.73 & 0.93 \\
          & 0.50 & 256.32 & 362.48 & 188.82 & 89.07 & 49.56 & 28.40 & 31.97 & 4.26 & 2.00 \\
          & 0.70 & 219.19 & 214.14 & 62.16 & 89.62 & 54.17 & 36.48 & 43.11 & 5.64 & 2.99 \\
          & 1.00 & 193.44 & 0.00 & -13.93 & 90.58 & 56.65 & 40.99 & 50.99 & 6.58 & 3.63 \\
\bottomrule
      \end{tabular}
    }
\end{table}

\begin{table}
  \caption{Dropout reconstruction with Guided Backprop on the Cora dataset using the GAT model.}
  \label{tab: GuidedBackprop-Cora-GAT}
  \centering
  \scalebox{0.7}{
      \begin{tabular}{ccccccccccc}
        \toprule
        \thead{Number \\ of clusters} &
        \thead{Dropout \\ rate} &
        \thead{Execution \\ time (s)} & 
        \thead{$\Delta$nodes \\ (\%)} & 
        \thead{$\Delta$edges \\ (\%)} &  
        \multicolumn{3}{c}{\thead{Affected \\ explanations (\%)}} &  
        \multicolumn{3}{c}{\thead{Affected \\ nodes (\%)}}  \\
        \cmidrule(lr){6-8}
        \cmidrule(lr){9-11}
        & & & & & 0.2 & 0.5 & 0.7 & 0.2 & 0.5 & 0.7 \\
        \midrule
        \multirow{1}{*}{1} 
          & 0.00 & 27.72 & 0.00 & 0.00 & 0.00 & 0.00 & 0.00 & 0.00 & 0.00 & 0.00 \\
        \midrule

        \multirow{5}{*}{16} 
          & 0.00 & 4.38 & 720.35 & 774.97 & 0.00 & 0.00 & 0.00 & 0.00 & 0.00 & 0.00 \\
          & 0.20 & 3.69 & 577.36 & 492.12 & 9.05 & 5.98 & 5.10 & 0.76 & 0.34 & 0.23 \\
          & 0.50 & 2.79 & 359.75 & 187.86 & 23.49 & 16.43 & 14.77 & 2.07 & 0.97 & 0.70 \\
          & 0.70 & 2.41 & 218.02 & 58.83 & 31.50 & 23.49 & 21.86 & 3.02 & 1.49 & 1.11 \\
          & 1.00 & 2.12 & 0.00 & -13.93 & 35.82 & 26.99 & 24.37 & 3.44 & 1.65 & 1.23 \\
\bottomrule
      \end{tabular}
    }
\end{table}

\begin{table}
  \caption{Dropout reconstruction with Saliency Map on the Cora dataset using the GAT model.}
  \label{tab: SaliencyMap-Cora-GAT}
  \centering
  \scalebox{0.7}{
      \begin{tabular}{ccccccccccc}
        \toprule
        \thead{Number \\ of clusters} &
        \thead{Dropout \\ rate} &
        \thead{Execution \\ time (s)} & 
        \thead{$\Delta$nodes \\ (\%)} & 
        \thead{$\Delta$edges \\ (\%)} &  
        \multicolumn{3}{c}{\thead{Affected \\ explanations (\%)}} &  
        \multicolumn{3}{c}{\thead{Affected \\ nodes (\%)}}  \\
        \cmidrule(lr){6-8}
        \cmidrule(lr){9-11}
        & & & & & 0.2 & 0.5 & 0.7 & 0.2 & 0.5 & 0.7 \\
        \midrule
        \multirow{1}{*}{1} 
          & 0.00 & 28.40 & 0.00 & 0.00 & 0.00 & 0.00 & 0.00 & 0.00 & 0.00 & 0.00 \\
        \midrule

        \multirow{5}{*}{16} 
          & 0.00 & 4.30 & 720.35 & 774.97 & 0.00 & 0.00 & 0.00 & 0.00 & 0.00 & 0.00 \\
          & 0.20 & 3.62 & 579.25 & 506.63 & 10.64 & 5.91 & 4.69 & 0.82 & 0.31 & 0.21 \\
          & 0.50 & 2.83 & 360.78 & 188.27 & 20.57 & 13.59 & 11.56 & 1.85 & 0.77 & 0.50 \\
          & 0.70 & 2.39 & 215.44 & 57.09 & 32.83 & 24.67 & 21.75 & 3.15 & 1.42 & 0.95 \\
          & 1.00 & 2.06 & 0.00 & -13.93 & 37.22 & 27.70 & 24.78 & 3.59 & 1.62 & 1.09 \\
\bottomrule
      \end{tabular}
    }
\end{table}


\begin{table}
  \caption{Dropout reconstruction with DeconvNet on the CiteSeer dataset using the GCN model.}
  \label{tab: Deconvnet-CiteSeer-GCN}
  \centering
  \scalebox{0.7}{
      \begin{tabular}{ccccccccccc}
        \toprule
        \thead{Number \\ of clusters} &
        \thead{Dropout \\ rate} &
        \thead{Execution \\ time (s)} & 
        \thead{$\Delta$nodes \\ (\%)} & 
        \thead{$\Delta$edges \\ (\%)} &  
        \multicolumn{3}{c}{\thead{Affected \\ explanations (\%)}} &  
        \multicolumn{3}{c}{\thead{Affected \\ nodes (\%)}}  \\
        \cmidrule(lr){6-8}
        \cmidrule(lr){9-11}
        & & & & & 0.2 & 0.5 & 0.7 & 0.2 & 0.5 & 0.7 \\
        \midrule
        \multirow{1}{*}{1} 
          & 0.00 & 21.56 & 0.00 & 0.00 & 0.00 & 0.00 & 0.00 & 0.00 & 0.00 & 0.00 \\
        \midrule

        \multirow{5}{*}{16} 
          & 0.00 & 2.40 & 144.27 & 216.10 & 0.00 & 0.00 & 0.00 & 0.00 & 0.00 & 0.00 \\
          & 0.20 & 2.24 & 115.33 & 133.48 & 3.88 & 0.51 & 0.21 & 0.42 & 0.03 & 0.01 \\
          & 0.50 & 1.93 & 72.86 & 54.57 & 7.69 & 1.20 & 0.36 & 0.97 & 0.08 & 0.02 \\
          & 0.70 & 1.80 & 43.70 & 15.55 & 10.31 & 1.50 & 0.51 & 1.39 & 0.11 & 0.03 \\
          & 1.00 & 1.67 & 0.00 & -6.72 & 11.72 & 2.31 & 0.81 & 1.68 & 0.17 & 0.05 \\
\bottomrule
      \end{tabular}
    }
\end{table}

\begin{table}
  \caption{Dropout reconstruction with GNNExplainer on the CiteSeer dataset using the GCN model.}
  \label{tab: GNNExplainer-CiteSeer-GCN}
  \centering
  \scalebox{0.7}{
      \begin{tabular}{ccccccccccc}
        \toprule
        \thead{Number \\ of clusters} &
        \thead{Dropout \\ rate} &
        \thead{Execution \\ time (s)} & 
        \thead{$\Delta$nodes \\ (\%)} & 
        \thead{$\Delta$edges \\ (\%)} &  
        \multicolumn{3}{c}{\thead{Affected \\ explanations (\%)}} &  
        \multicolumn{3}{c}{\thead{Affected \\ nodes (\%)}}  \\
        \cmidrule(lr){6-8}
        \cmidrule(lr){9-11}
        & & & & & 0.2 & 0.5 & 0.7 & 0.2 & 0.5 & 0.7 \\
        \midrule
        \multirow{1}{*}{1} 
          & 0.00 & 2340.40 & 0.00 & 0.00 & 0.00 & 0.00 & 0.00 & 0.00 & 0.00 & 0.00 \\
        \midrule

        \multirow{5}{*}{16} 
          & 0.00 & 211.36 & 144.27 & 216.10 & 72.50 & 41.03 & 0.00 & 18.96 & 9.07 & 0.00 \\
          & 0.20 & 190.66 & 114.55 & 138.44 & 74.66 & 42.20 & 7.27 & 25.03 & 9.95 & 1.09 \\
          & 0.50 & 169.88 & 72.20 & 51.80 & 75.08 & 43.13 & 10.34 & 32.68 & 11.15 & 2.22 \\
          & 0.70 & 156.77 & 42.35 & 13.42 & 74.96 & 42.17 & 13.04 & 38.60 & 11.62 & 3.12 \\
          & 1.00 & 148.35 & 0.00 & -6.72 & 74.78 & 41.87 & 14.76 & 41.97 & 12.08 & 3.80 \\
\bottomrule
      \end{tabular}
    }
\end{table}

\begin{table}
  \caption{Dropout reconstruction with Guided Backprop on the CiteSeer dataset using the GCN model.}
  \label{tab: GuidedBackprop-CiteSeer-GCN}
  \centering
  \scalebox{0.7}{
      \begin{tabular}{ccccccccccc}
        \toprule
        \thead{Number \\ of clusters} &
        \thead{Dropout \\ rate} &
        \thead{Execution \\ time (s)} & 
        \thead{$\Delta$nodes \\ (\%)} & 
        \thead{$\Delta$edges \\ (\%)} &  
        \multicolumn{3}{c}{\thead{Affected \\ explanations (\%)}} &  
        \multicolumn{3}{c}{\thead{Affected \\ nodes (\%)}}  \\
        \cmidrule(lr){6-8}
        \cmidrule(lr){9-11}
        & & & & & 0.2 & 0.5 & 0.7 & 0.2 & 0.5 & 0.7 \\
        \midrule
        \multirow{1}{*}{1} 
          & 0.00 & 21.74 & 0.00 & 0.00 & 0.00 & 0.00 & 0.00 & 0.00 & 0.00 & 0.00 \\
        \midrule

        \multirow{5}{*}{16} 
          & 0.00 & 2.41 & 144.27 & 216.10 & 0.00 & 0.00 & 0.00 & 0.00 & 0.00 & 0.00 \\
          & 0.20 & 2.26 & 115.87 & 142.42 & 4.39 & 0.57 & 0.24 & 0.47 & 0.04 & 0.02 \\
          & 0.50 & 1.94 & 72.35 & 51.34 & 8.21 & 1.14 & 0.33 & 1.03 & 0.08 & 0.02 \\
          & 0.70 & 1.81 & 42.41 & 16.83 & 9.53 & 1.68 & 0.57 & 1.38 & 0.12 & 0.04 \\
          & 1.00 & 1.69 & 0.00 & -6.72 & 11.72 & 2.31 & 0.81 & 1.68 & 0.17 & 0.05 \\
\bottomrule
      \end{tabular}
    }
\end{table}

\begin{table}
  \caption{Dropout reconstruction with Saliency Map on the CiteSeer dataset using the GCN model.}
  \label{tab: SaliencyMap-CiteSeer-GCN}      
  \centering
  \scalebox{0.7}{
      \begin{tabular}{ccccccccccc}
        \toprule
        \thead{Number \\ of clusters} &
        \thead{Dropout \\ rate} &
        \thead{Execution \\ time (s)} & 
        \thead{$\Delta$nodes \\ (\%)} & 
        \thead{$\Delta$edges \\ (\%)} &  
        \multicolumn{3}{c}{\thead{Affected \\ explanations (\%)}} &  
        \multicolumn{3}{c}{\thead{Affected \\ nodes (\%)}}  \\
        \cmidrule(lr){6-8}
        \cmidrule(lr){9-11}
        & & & & & 0.2 & 0.5 & 0.7 & 0.2 & 0.5 & 0.7 \\
        \midrule
        \multirow{1}{*}{1} 
          & 0.00 & 21.96 & 0.00 & 0.00 & 0.00 & 0.00 & 0.00 & 0.00 & 0.00 & 0.00 \\
        \midrule

        \multirow{5}{*}{16} 
          & 0.00 & 2.39 & 144.27 & 216.10 & 0.00 & 0.00 & 0.00 & 0.00 & 0.00 & 0.00 \\
          & 0.20 & 2.18 & 115.57 & 137.85 & 3.61 & 0.36 & 0.12 & 0.37 & 0.03 & 0.01 \\
          & 0.50 & 1.96 & 70.33 & 48.57 & 8.51 & 1.23 & 0.36 & 1.07 & 0.08 & 0.02 \\
          & 0.70 & 1.80 & 41.33 & 14.96 & 10.31 & 1.62 & 0.51 & 1.41 & 0.11 & 0.03 \\
          & 1.00 & 1.64 & 0.00 & -6.72 & 11.75 & 2.31 & 0.81 & 1.68 & 0.17 & 0.05 \\
\bottomrule
      \end{tabular}
    }
\end{table}

\begin{table}
  \caption{Dropout reconstruction with DeconvNet on the CiteSeer dataset using the GAT model.}
  \label{tab: Deconvnet-CiteSeer-GAT}      
  \centering
  \scalebox{0.7}{
      \begin{tabular}{ccccccccccc}
        \toprule
        \thead{Number \\ of clusters} &
        \thead{Dropout \\ rate} &
        \thead{Execution \\ time (s)} & 
        \thead{$\Delta$nodes \\ (\%)} & 
        \thead{$\Delta$edges \\ (\%)} &  
        \multicolumn{3}{c}{\thead{Affected \\ explanations (\%)}} &  
        \multicolumn{3}{c}{\thead{Affected \\ nodes (\%)}}  \\
        \cmidrule(lr){6-8}
        \cmidrule(lr){9-11}
        & & & & & 0.2 & 0.5 & 0.7 & 0.2 & 0.5 & 0.7 \\
        \midrule
        \multirow{1}{*}{1} 
          & 0.00 & 35.61 & 0.00 & 0.00 & 0.00 & 0.00 & 0.00 & 0.00 & 0.00 & 0.00 \\
        \midrule

        \multirow{5}{*}{16} 
          & 0.00 & 3.84 & 144.27 & 216.10 & 0.00 & 0.00 & 0.00 & 0.00 & 0.00 & 0.00 \\
          & 0.20 & 3.55 & 115.09 & 132.07 & 4.78 & 3.49 & 2.98 & 1.14 & 0.56 & 0.41 \\
          & 0.50 & 3.24 & 70.24 & 50.92 & 10.52 & 7.60 & 6.70 & 2.32 & 1.16 & 0.86 \\
          & 0.70 & 3.05 & 44.88 & 16.72 & 13.07 & 9.41 & 8.66 & 3.09 & 1.53 & 1.14 \\
          & 1.00 & 2.93 & 0.00 & -6.72 & 15.06 & 10.79 & 9.59 & 3.46 & 1.72 & 1.27 \\
\bottomrule
      \end{tabular}
    }
\end{table}

\begin{table}
  \caption{Dropout reconstruction with GNNExplainer on the CiteSeer dataset using the GAT model.}
  \label{tab: GNNExplainer-CiteSeer-GAT}      
  \centering
  \scalebox{0.7}{
      \begin{tabular}{ccccccccccc}
        \toprule
        \thead{Number \\ of clusters} &
        \thead{Dropout \\ rate} &
        \thead{Execution \\ time (s)} & 
        \thead{$\Delta$nodes \\ (\%)} & 
        \thead{$\Delta$edges \\ (\%)} &  
        \multicolumn{3}{c}{\thead{Affected \\ explanations (\%)}} &  
        \multicolumn{3}{c}{\thead{Affected \\ nodes (\%)}}  \\
        \cmidrule(lr){6-8}
        \cmidrule(lr){9-11}
        & & & & & 0.2 & 0.5 & 0.7 & 0.2 & 0.5 & 0.7 \\
        \midrule
        \multirow{1}{*}{1} 
          & 0.00 & 3816.54 & 0.00 & 0.00 & 0.00 & 0.00 & 0.00 & 0.00 & 0.00 & 0.00 \\
        \midrule

        \multirow{5}{*}{16} 
          & 0.00 & 325.03 & 144.27 & 216.10 & 66.55 & 26.57 & 0.00 & 13.86 & 3.45 & 0.00 \\
          & 0.20 & 302.25 & 116.56 & 142.16 & 70.33 & 30.81 & 7.78 & 23.21 & 4.68 & 0.82 \\
          & 0.50 & 271.80 & 72.02 & 50.33 & 71.84 & 32.55 & 12.05 & 28.15 & 6.08 & 1.80 \\
          & 0.70 & 254.67 & 43.31 & 15.42 & 71.84 & 33.78 & 15.90 & 34.05 & 6.75 & 2.51 \\
          & 1.00 & 241.96 & 0.00 & -6.72 & 72.71 & 34.45 & 17.22 & 38.26 & 7.59 & 3.06 \\
\bottomrule
      \end{tabular}
    }
\end{table}

\begin{table}
  \caption{Dropout reconstruction with Guided Backprop on the CiteSeer dataset using the GAT model.}
  \label{tab: GuidedBackprop-CiteSeer-GAT}      
  \centering
  \scalebox{0.7}{
      \begin{tabular}{ccccccccccc}
        \toprule
        \thead{Number \\ of clusters} &
        \thead{Dropout \\ rate} &
        \thead{Execution \\ time (s)} & 
        \thead{$\Delta$nodes \\ (\%)} & 
        \thead{$\Delta$edges \\ (\%)} &  
        \multicolumn{3}{c}{\thead{Affected \\ explanations (\%)}} &  
        \multicolumn{3}{c}{\thead{Affected \\ nodes (\%)}}  \\
        \cmidrule(lr){6-8}
        \cmidrule(lr){9-11}
        & & & & & 0.2 & 0.5 & 0.7 & 0.2 & 0.5 & 0.7 \\
        \midrule
        \multirow{1}{*}{1} 
          & 0.00 & 36.03 & 0.00 & 0.00 & 0.00 & 0.00 & 0.00 & 0.00 & 0.00 & 0.00 \\
        \midrule

        \multirow{5}{*}{16} 
          & 0.00 & 3.53 & 144.27 & 216.10 & 0.00 & 0.00 & 0.00 & 0.00 & 0.00 & 0.00 \\
          & 0.20 & 3.30 & 114.07 & 135.90 & 3.40 & 2.22 & 2.04 & 0.80 & 0.39 & 0.28 \\
          & 0.50 & 3.03 & 72.77 & 51.41 & 11.51 & 8.54 & 7.54 & 2.65 & 1.32 & 0.96 \\
          & 0.70 & 2.97 & 43.37 & 17.60 & 9.53 & 5.98 & 5.02 & 1.92 & 0.87 & 0.63 \\
          & 1.00 & 2.86 & 0.00 & -6.72 & 15.06 & 10.76 & 9.56 & 3.46 & 1.72 & 1.27 \\
\bottomrule
      \end{tabular}
    }
\end{table}

\begin{table}
  \caption{Dropout reconstruction with Saliency Map on the CiteSeer dataset using the GAT model.}
  \label{tab: SaliencyMap-CiteSeer-GAT}      
  \centering
  \scalebox{0.7}{
      \begin{tabular}{ccccccccccc}
        \toprule
        \thead{Number \\ of clusters} &
        \thead{Dropout \\ rate} &
        \thead{Execution \\ time (s)} & 
        \thead{$\Delta$nodes \\ (\%)} & 
        \thead{$\Delta$edges \\ (\%)} &  
        \multicolumn{3}{c}{\thead{Affected \\ explanations (\%)}} &  
        \multicolumn{3}{c}{\thead{Affected \\ nodes (\%)}}  \\
        \cmidrule(lr){6-8}
        \cmidrule(lr){9-11}
        & & & & & 0.2 & 0.5 & 0.7 & 0.2 & 0.5 & 0.7 \\
        \midrule
        \multirow{1}{*}{1} 
          & 0.00 & 35.65 & 0.00 & 0.00 & 0.00 & 0.00 & 0.00 & 0.00 & 0.00 & 0.00 \\
        \midrule

        \multirow{5}{*}{16} 
          & 0.00 & 3.57 & 144.27 & 216.10 & 0.00 & 0.00 & 0.00 & 0.00 & 0.00 & 0.00 \\
          & 0.20 & 3.28 & 116.95 & 141.10 & 2.74 & 1.41 & 1.08 & 0.49 & 0.17 & 0.11 \\
          & 0.50 & 2.95 & 71.51 & 48.68 & 9.86 & 6.55 & 5.80 & 2.17 & 1.00 & 0.73 \\
          & 0.70 & 2.77 & 44.63 & 18.26 & 13.23 & 9.35 & 8.24 & 2.93 & 1.43 & 1.03 \\
          & 1.00 & 2.62 & 0.00 & -6.72 & 15.21 & 10.88 & 9.65 & 3.45 & 1.69 & 1.26 \\
\bottomrule
      \end{tabular}
    }
\end{table}


\begin{table}
  \caption{Dropout reconstruction with DeconvNet on the PubMed dataset using the GCN model.}
  \label{tab: Deconvnet-PubMed-GCN}      
  \centering
  \scalebox{0.7}{
      \begin{tabular}{ccccccccccc}
        \toprule
        \thead{Number \\ of clusters} &
        \thead{Dropout \\ rate} &
        \thead{Execution \\ time (s)} & 
        \thead{$\Delta$nodes \\ (\%)} & 
        \thead{$\Delta$edges \\ (\%)} &  
        \multicolumn{3}{c}{\thead{Affected \\ explanations (\%)}} &  
        \multicolumn{3}{c}{\thead{Affected \\ nodes (\%)}}  \\
        \cmidrule(lr){6-8}
        \cmidrule(lr){9-11}
        & & & & & 0.2 & 0.5 & 0.7 & 0.2 & 0.5 & 0.7 \\
        \midrule
        \multirow{1}{*}{1} 
          & 0.00 & 283.01 & 0.00 & 0.00 & 0.00 & 0.00 & 0.00 & 0.00 & 0.00 & 0.00 \\
        \midrule
        
        \multirow{5}{*}{16} 
          & 0.00 & 133.09 & 875.45 & 1020.18 & 0.00 & 0.00 & 0.00 & 0.00 & 0.00 & 0.00 \\
          & 0.20 & 96.07 & 701.05 & 654.14 & 9.91 & 0.47 & 0.07 & 0.35 & 0.01 & 0.00 \\
          & 0.50 & 53.63 & 436.46 & 252.70 & 17.68 & 1.20 & 0.19 & 0.81 & 0.02 & 0.00 \\
          & 0.70 & 34.68 & 261.85 & 82.38 & 21.03 & 1.87 & 0.23 & 1.11 & 0.04 & 0.00 \\
          & 1.00 & 17.80 & 0.00 & -16.70 & 24.19 & 2.27 & 0.34 & 1.26 & 0.04 & 0.01 \\
        \bottomrule
      \end{tabular}
    }
\end{table}

\begin{table}
  \caption{Dropout reconstruction with Guided Backprop on the PubMed dataset using the GCN model.}
  \label{tab: GuidedBackprop-PubMed-GCN}      
  \centering
  \scalebox{0.7}{
      \begin{tabular}{ccccccccccc}
        \toprule
        \thead{Number \\ of clusters} &
        \thead{Dropout \\ rate} &
        \thead{Execution \\ time (s)} & 
        \thead{$\Delta$nodes \\ (\%)} & 
        \thead{$\Delta$edges \\ (\%)} &  
        \multicolumn{3}{c}{\thead{Affected \\ explanations (\%)}} &  
        \multicolumn{3}{c}{\thead{Affected \\ nodes (\%)}}  \\
        \cmidrule(lr){6-8}
        \cmidrule(lr){9-11}
        & & & & & 0.2 & 0.5 & 0.7 & 0.2 & 0.5 & 0.7 \\
        \midrule
        \multirow{1}{*}{1} 
          & 0.00 & 283.40 & 0.00 & 0.00 & 0.00 & 0.00 & 0.00 & 0.00 & 0.00 & 0.00 \\
        \midrule
        
        \multirow{5}{*}{16} 
          & 0.00 & 133.50 & 875.45 & 1020.18 & 0.00 & 0.00 & 0.00 & 0.00 & 0.00 & 0.00 \\
          & 0.20 & 96.65 & 700.08 & 651.77 & 9.63 & 0.55 & 0.08 & 0.34 & 0.01 & 0.00 \\
          & 0.50 & 54.00 & 437.61 & 251.60 & 17.19 & 1.35 & 0.16 & 0.79 & 0.03 & 0.00 \\
          & 0.70 & 34.75 & 261.46 & 84.25 & 21.56 & 1.83 & 0.25 & 1.12 & 0.03 & 0.00 \\
          & 1.00 & 18.26 & 0.00 & -16.70 & 24.20 & 2.28 & 0.34 & 1.26 & 0.04 & 0.01 \\
        \bottomrule
      \end{tabular}
    }
\end{table}

\begin{table}
  \caption{Dropout reconstruction with Saliency Map on the PubMed dataset using the GCN model}
  \label{tab: SaliencyMap-PubMed-GCN}      
  \centering
  \scalebox{0.7}{
      \begin{tabular}{ccccccccccc}
        \toprule
        \thead{Number \\ of clusters} &
        \thead{Dropout \\ rate} &
        \thead{Execution \\ time (s)} & 
        \thead{$\Delta$nodes \\ (\%)} & 
        \thead{$\Delta$edges \\ (\%)} &  
        \multicolumn{3}{c}{\thead{Affected \\ explanations (\%)}} &  
        \multicolumn{3}{c}{\thead{Affected \\ nodes (\%)}}  \\
        \cmidrule(lr){6-8}
        \cmidrule(lr){9-11}
        & & & & & 0.2 & 0.5 & 0.7 & 0.2 & 0.5 & 0.7 \\
        \midrule
        \multirow{1}{*}{1} 
          & 0.00 & 294.67 & 0.00 & 0.00 & 0.00 & 0.00 & 0.00 & 0.00 & 0.00 & 0.00 \\
        \midrule
        
        \multirow{5}{*}{16} 
          & 0.00 & 132.81 & 875.45 & 1020.18 & 0.00 & 0.00 & 0.00 & 0.00 & 0.00 & 0.00 \\
          & 0.20 & 95.81 & 699.60 & 650.67 & 9.60 & 0.50 & 0.07 & 0.32 & 0.01 & 0.00 \\
          & 0.50 & 53.65 & 438.81 & 248.93 & 16.91 & 1.32 & 0.20 & 0.76 & 0.02 & 0.00 \\
          & 0.70 & 34.38 & 261.90 & 83.51 & 21.51 & 1.97 & 0.30 & 1.09 & 0.04 & 0.01 \\
          & 1.00 & 17.79 & 0.00 & -16.70 & 24.05 & 2.42 & 0.39 & 1.22 & 0.05 & 0.01 \\
        \bottomrule
      \end{tabular}
    }
\end{table}

\begin{table}
  \caption{Dropout reconstruction with DeconvNet on the PubMed dataset using the GAT model.}
  \label{tab: Deconvnet-PubMed-GAT}      
  \centering
  \scalebox{0.7}{
      \begin{tabular}{ccccccccccc}
        \toprule
        \thead{Number \\ of clusters} &
        \thead{Dropout \\ rate} &
        \thead{Execution \\ time (s)} & 
        \thead{$\Delta$nodes \\ (\%)} & 
        \thead{$\Delta$edges \\ (\%)} &  
        \multicolumn{3}{c}{\thead{Affected \\ explanations (\%)}} &  
        \multicolumn{3}{c}{\thead{Affected \\ nodes (\%)}}  \\
        \cmidrule(lr){6-8}
        \cmidrule(lr){9-11}
        & & & & & 0.2 & 0.5 & 0.7 & 0.2 & 0.5 & 0.7 \\
        \midrule
        \multirow{1}{*}{1} 
          & 0.00 & 403.32 & 0.00 & 0.00 & 0.00 & 0.00 & 0.00 & 0.00 & 0.00 & 0.00 \\
        \midrule
        
        \multirow{5}{*}{16} 
          & 0.00 & 182.39 & 875.45 & 1020.18 & 0.00 & 0.00 & 0.00 & 0.00 & 0.00 & 0.00 \\
          & 0.20 & 132.23 & 700.79 & 648.32 & 16.07 & 6.44 & 4.13 & 0.68 & 0.18 & 0.09 \\
          & 0.50 & 75.30 & 438.18 & 251.91 & 29.48 & 13.71 & 9.59 & 1.65 & 0.43 & 0.23 \\
          & 0.70 & 48.17 & 262.44 & 82.84 & 35.72 & 18.18 & 13.00 & 2.22 & 0.60 & 0.33 \\
          & 1.00 & 25.03 & 0.00 & -16.70 & 41.61 & 21.70 & 16.18 & 2.78 & 0.76 & 0.42 \\
        \bottomrule
      \end{tabular}
    }
\end{table}

\begin{table}
  \caption{Dropout reconstruction with Guided Backprop on the PubMed dataset using the GAT model.}
  \label{tab: GuidedBackprop-PubMed-GAT}      
  \centering
  \scalebox{0.7}{
      \begin{tabular}{ccccccccccc}
        \toprule
        \thead{Number \\ of clusters} &
        \thead{Dropout \\ rate} &
        \thead{Execution \\ time (s)} & 
        \thead{$\Delta$nodes \\ (\%)} & 
        \thead{$\Delta$edges \\ (\%)} &  
        \multicolumn{3}{c}{\thead{Affected \\ explanations (\%)}} &  
        \multicolumn{3}{c}{\thead{Affected \\ nodes (\%)}}  \\
        \cmidrule(lr){6-8}
        \cmidrule(lr){9-11}
        & & & & & 0.2 & 0.5 & 0.7 & 0.2 & 0.5 & 0.7 \\
        \midrule
        \multirow{1}{*}{1} 
          & 0.00 & 399.74 & 0.00 & 0.00 & 0.00 & 0.00 & 0.00 & 0.00 & 0.00 & 0.00 \\
        \midrule
        
        \multirow{5}{*}{16} 
          & 0.00 & 182.89 & 875.45 & 1020.18 & 0.00 & 0.00 & 0.00 & 0.00 & 0.00 & 0.00 \\
          & 0.20 & 132.46 & 700.33 & 651.08 & 16.07 & 6.33 & 4.12 & 0.72 & 0.18 & 0.10 \\
          & 0.50 & 75.51 & 437.71 & 252.07 & 29.30 & 13.67 & 9.71 & 1.68 & 0.44 & 0.24 \\
          & 0.70 & 48.27 & 264.23 & 85.06 & 34.61 & 17.41 & 12.47 & 2.13 & 0.58 & 0.31 \\
          & 1.00 & 25.64 & 0.00 & -16.70 & 41.61 & 21.70 & 16.18 & 2.78 & 0.76 & 0.42 \\
        \bottomrule
      \end{tabular}
    }
\end{table}

\begin{table}
  \caption{Dropout reconstruction with Saliency Map on the PubMed dataset using the GAT model.}
  \label{tab: SaliencyMap-PubMed-GAT}  
  \centering
  \scalebox{0.7}{
      \begin{tabular}{ccccccccccc}
        \toprule
        \thead{Number \\ of clusters} &
        \thead{Dropout \\ rate} &
        \thead{Execution \\ time (s)} & 
        \thead{$\Delta$nodes \\ (\%)} & 
        \thead{$\Delta$edges \\ (\%)} &  
        \multicolumn{3}{c}{\thead{Affected \\ explanations (\%)}} &  
        \multicolumn{3}{c}{\thead{Affected \\ nodes (\%)}}  \\
        \cmidrule(lr){6-8}
        \cmidrule(lr){9-11}
        & & & & & 0.2 & 0.5 & 0.7 & 0.2 & 0.5 & 0.7 \\
        \midrule
        \multirow{1}{*}{1} 
          & 0.00 & 401.21 & 0.00 & 0.00 & 0.00 & 0.00 & 0.00 & 0.00 & 0.00 & 0.00 \\
        \midrule
        
        \multirow{5}{*}{16} 
          & 0.00 & 182.07 & 875.45 & 1020.18 & 0.00 & 0.00 & 0.00 & 0.00 & 0.00 & 0.00 \\
          & 0.20 & 132.18 & 701.16 & 653.18 & 15.97 & 6.31 & 4.22 & 0.71 & 0.19 & 0.11 \\
          & 0.50 & 75.33 & 440.30 & 250.29 & 30.67 & 14.18 & 9.93 & 1.71 & 0.45 & 0.24 \\
          & 0.70 & 48.14 & 263.62 & 83.85 & 35.62 & 17.55 & 12.81 & 2.19 & 0.59 & 0.33 \\
          & 1.00 & 25.10 & 0.00 & -16.70 & 41.65 & 21.61 & 16.13 & 2.78 & 0.76 & 0.42 \\
        \bottomrule
      \end{tabular}
    }
\end{table}

\end{document}